\documentclass[runningheads]{llncs}

\usepackage{eccv}
\usepackage{eccvabbrv}

\usepackage{graphicx}
\usepackage{booktabs}

\usepackage[accsupp]{axessibility}

\usepackage[hidelinks]{hyperref}

\usepackage{orcidlink}

\usepackage[table]{xcolor}      
\usepackage{makecell}
\usepackage{pifont}

\definecolor{navyblue}{HTML}{0071BC}
\definecolor{hotpink}{HTML}{FF0080}
\definecolor{oai-white}{HTML}{FFFFFF}
\definecolor{oai-black}{HTML}{000000}
\definecolor{oai-red}{HTML}{FF4500}
\definecolor{oai-green}{HTML}{51DA4C}
\definecolor{oai-blue}{HTML}{0000FF}
\definecolor{oai-yellow}{HTML}{FFF639}
\definecolor{oai-magenta}{HTML}{FF45FF}
\definecolor{oai-cyan}{HTML}{00FFFF}
\definecolor{oai-orange}{HTML}{FE7600}
\definecolor{oai-violet}{HTML}{8A2BE2}
\definecolor{oai-brown}{HTML}{A0522D}
\definecolor{oai-green-050}{HTML}{F4FFF4}
\definecolor{oai-green-100}{HTML}{E9FFE8}
\definecolor{oai-green-200}{HTML}{D9FFD8}
\definecolor{oai-green-300}{HTML}{C9FFC7}
\definecolor{oai-green-400}{HTML}{A6FFA3}
\definecolor{oai-green-500}{HTML}{7CF178}
\definecolor{oai-green-600}{HTML}{51DA4C}
\definecolor{oai-green-700}{HTML}{3FA93B}
\definecolor{oai-green-800}{HTML}{2D712A}
\definecolor{oai-green-900}{HTML}{193718}
\definecolor{oai-gray-000}{HTML}{FFFFFF}
\definecolor{oai-gray-100}{HTML}{FAFAFA}
\definecolor{oai-gray-200}{HTML}{F5F5F5}
\definecolor{oai-gray-300}{HTML}{E5E5E5}
\definecolor{oai-gray-400}{HTML}{FFB7A4}
\definecolor{oai-gray-500}{HTML}{CDCDCD}
\definecolor{oai-gray-600}{HTML}{A8A8A8}
\definecolor{oai-gray-700}{HTML}{747474}
\definecolor{oai-gray-800}{HTML}{393939}
\definecolor{oai-gray-900}{HTML}{000000}
\usepackage{algorithm}
\usepackage{algpseudocode} 
\algrenewcommand\algorithmicrequire{\textbf{Input}}
\algrenewcommand\algorithmicensure{\textbf{Output}}
\usepackage{xcolor} 
\usepackage[most]{tcolorbox}
\usepackage{multirow}
\usepackage{graphicx}
\usepackage[table]{xcolor}
\definecolor{myblue}{RGB}{230,245,255}
\definecolor{mygreen}{HTML}{FFD500}
\definecolor{navyblue}{HTML}{0071BC}
\definecolor{darkgreen}{rgb}{0.0, 0.5, 0.0}
\newcommand{\MethodName}{VLM\textsuperscript{2}}

\usepackage{enumitem}

\let\llncssubparagraph\subparagraph
\let\subparagraph\paragraph
\usepackage[compact]{titlesec}
\let\subparagraph\llncssubparagraph
\titlespacing{\section}{0pt}{3ex}{2ex}
\titlespacing{\subsection}{0pt}{1.5ex}{1ex}

\begin{document}

\title{Vision-Language Memory for Spatial Reasoning} 

\author{Zuntao Liu\inst{1}\orcidlink{0009-0004-1351-6847} \and
Yi Du\inst{1}\orcidlink{0009-0007-4312-2170} \and
Taimeng Fu\inst{1}\orcidlink{0000-0003-4545-7736} \and
Shaoshu Su\inst{1}\orcidlink{0000-0003-1690-2327} \and
Cherie Ho\inst{2}\orcidlink{0000-0003-1886-1020} \and
Chen Wang\inst{1}\thanks{Corresponding author.}\orcidlink{0000-0002-4630-0805}
}

\authorrunning{Z.~Liu et al.}

\institute{Spatial AI \& Robotics (SAIR) Lab, University at Buffalo, USA\\
\and
Stanford University, USA\\
\email{lzuntao327@gmail.com, chenw@sairlab.org}\\
\url{https://sairlab.org/vlm2}}

\maketitle

\begin{abstract}
Spatial reasoning is a critical capability for intelligent robots, yet current vision-language models (VLMs) still fall short of human-level performance in video-based spatial reasoning. This gap mainly stems from two challenges: a semantic-geometric misalignment that prevents consistent 3D understanding, and the absence of persistent memory to retain 3D representation and understanding across frames. 
To address these limitations, we present VLM$^2$, a Vision-Language Model with persistent Memory for spatial reasoning with a view-consistent, 3D-aware representation purely from 2D videos.
Specifically, we incorporate a dual-memory module consisting of a working memory that operates as a sliding window to focus on immediate context, and an episodic memory that consolidates and stores critical information across frames. 
This design enables bounded and efficient spatial reasoning under a fixed computational cost. Extensive experiments on multiple benchmarks show that VLM$^2$ achieves state-of-the-art performance among video-based models, significantly advancing the frontier of visual-spatial intelligence. 
\end{abstract}

\section{Introduction}
\begin{figure}[t]
\centering
\includegraphics[width=\columnwidth]{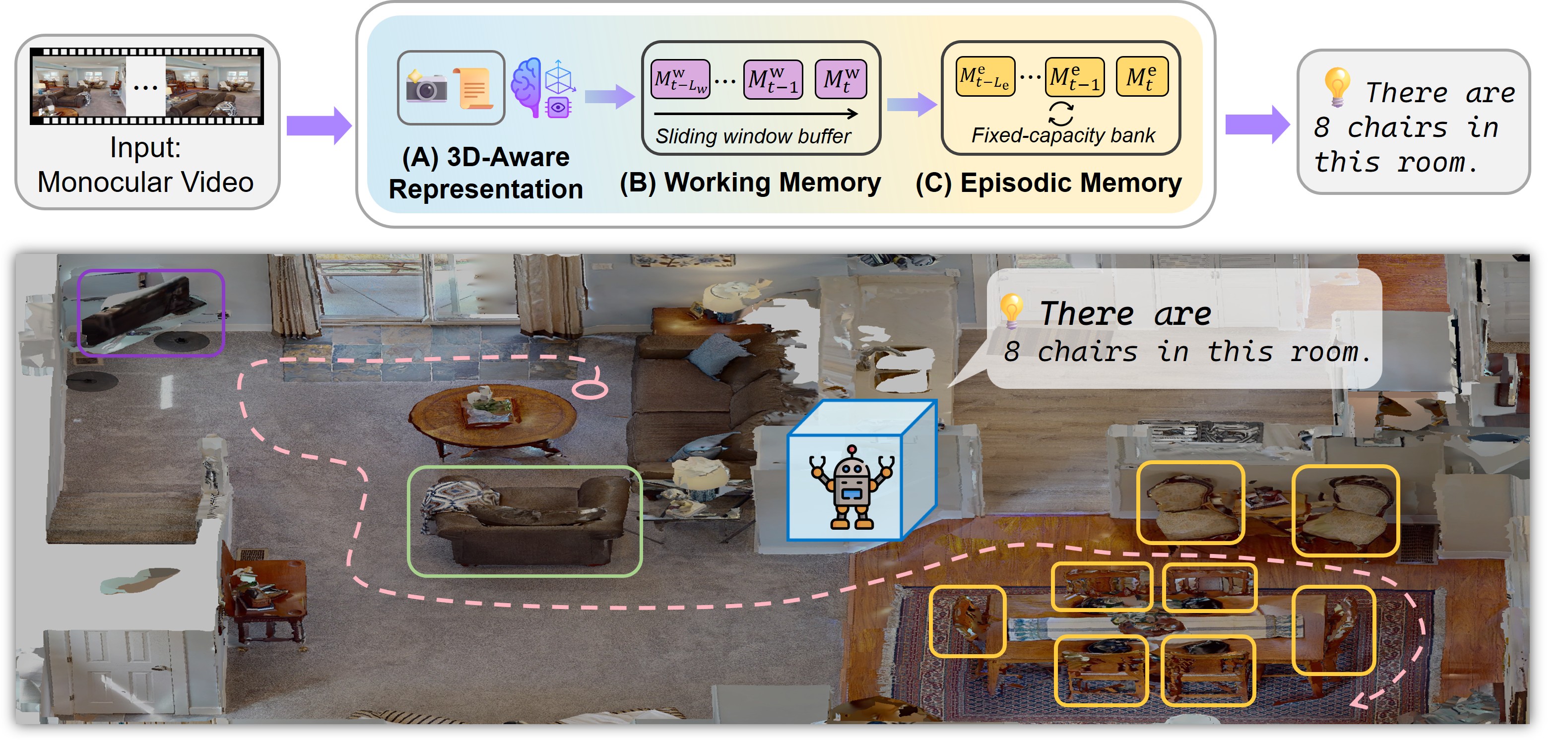}
\caption{\textbf{\MethodName{} is a \underline{V}ision-\underline{L}anguage \underline{M}odel with \underline{M}emory for spatial reasoning} that constructs view-consistent 3D-aware representations from monocular video and supports consistent reasoning across viewpoints. Such capabilities are critical for questions like ``How many chairs are in this room?'', which require consistent cross-view alignment and reasoning over previously observed content.
}
\label{fig:figintro}
\end{figure}

Spatial reasoning is a fundamental capability of intelligent robots, enabling them to perceive, localize, and reason about spatial relationships in the physical world. Recent advances~\cite{huang2023chat,chen2024ll3da,deng20253d,zheng2025video,wang2025ross3d,liu2025ssr} have sought to enhance this ability in vision-language models (VLMs)~\cite{bai2025qwen2,hurst2024gpt4o,lin2024vila,li2025llavaonevision,llavanext}, for instance by incorporating 3D information such as point clouds and depth maps to enhance spatial awareness. Despite these efforts, current VLMs still fall short of human-level spatial reasoning~\cite{yang2025thinking,chen2024spatialvlm}.
As in Fig.~\ref{fig:figintro}, existing models often fail to answer simple questions like \textit{``How many chairs are there in this room?''}. One reason is that such questions require consistent feature alignment across multiple views (recognizing the same chairs across viewpoints) and memory across frames (maintaining an accurate count as views change).
However, existing VLMs rarely address the two issues simultaneously.
In practice, they often exhibit (1) semantic-geometric feature misalignment and (2) the absence of persistent memory.

First, while existing methods that leverage semantic features from 2D encoders provide strong categorical understanding, they lack metric grounding and positional awareness~\cite{yuan2025empowering,huang2025mllms,zheng2025video,diomataris2021grounding}, making it challenging to maintain spatial coherence across viewpoints.
For instance, when moving from the living room to the kitchen, a model may incorrectly associate distinct chairs at different locations as the same instance, causing spatial inconsistency. 
In contrast, geometric features from 3D visual geometry models offer reliable structural cues; 
however, naïvely fusing them with semantic representations often produces view-dependent instabilities~\cite{liao2024ov,liao2025clip,cen2025tackling,shi20243DOVS}, causing global spatial inconsistencies under camera motion.

Moreover, directly using predicted 3D point clouds as explicit inputs does not inherently establish semantic-geometric correspondence across views. Although they provide structural cues, point-level predictions are not naturally aligned with semantic tokens, thus requiring additional alignment mechanisms.

The second challenge lies in the absence of persistent memory, although memory mechanisms have been explored in computer vision for decades \cite{hochreiter1997long}. 
Mobile robots experience viewpoint changes and occlusions, where objects may temporarily disappear and reappear as the viewpoint shifts~\cite{xu2020learning,lecun2022path}. 
Existing VLMs~\cite{llavanext,bai2025qwen2,li2025llavaonevision,openai2024gpt4o} rely on token-based context windows that struggle to preserve coherent spatial representations across frames~\cite{yang20253dmem,hu20253dllmmem}. 
As new visual inputs arrive, previously observed information may be overwritten, preventing consistent spatial reasoning across the video sequence.
For example, when the viewpoint shifts and later revisits a previously observed area, a chair initially seen at the entrance may no longer be recalled, leading to failures in tasks such as object counting. 
This highlights the need for an explicit memory mechanism capable of preserving and updating 3D-aware representations across frames and viewpoints.

To solve these challenges, existing approaches have explored several directions. Early works~\cite{huang2023chat,chen2024ll3da,deng20253d,zheng2025video,wang2025ross3d,liu2025ssr} fine-tuned VLMs with additional 3D data, yet progress was constrained by limited dataset scale and diversity. More recent methods~\cite{wu2025spatialmllm,fan2025vlm3r,zheng2025vgllm} leverage geometric priors from 3D visual geometry foundation models~\cite{wang2024dust3r,wang2025vggt,wang2025cut3r}, but their reliance on simple feature fusion neither explicitly addresses semantic-geometric misalignment nor provides a dedicated mechanism to preserve representations across frames.
As a result, achieving coherent 3D-aware representations across views remains challenging.

In this paper, we present \MethodName{}, a \textbf{V}ision-\textbf{L}anguage \textbf{M}odel with persistent attention-based \textbf{M}emory for spatial reasoning.
First, we construct a 3D-aware representation from visual observations by enforcing the alignment between the semantic and geometric features before fusion. 
To this end, we introduce a viewpoint-aware geometry alignment module to align geometry tokens with their view tokens, ensuring geometric features from different viewpoints are distinct even if their geometry is similar.
Furthermore, we inject the predicted 3D coordinates into their corresponding visual tokens to ensure geometric awareness.
To avoid potential spatial inconsistency, we introduce a learnable gate to decide which of these 3D points are useful, allowing the model to be ``adaptive.''
These designs unlock the model's ability to perform spatial reasoning, yielding view-consistent 3D-aware representations under camera motion.

Additionally, some models like~\cite{zheng2025video} require both 2D videos and accurate depth labels (or 3D point clouds) as inputs. This requirement limits their applicability in real-world scenarios, where reliable depth measurements are often unavailable. Although depth maps can be estimated using monocular depth estimation models, our empirical observations show that directly using such noisy depth predictions can significantly degrade model performance.
In contrast, our experiments demonstrate that our proposed view-consistent 3D-aware representation effectively addresses these issues without relying on accurate depth inputs.

To equip the model with persistent memory, we introduce an attention-based dual-memory to preserve and update 3D-aware representations across frames and viewpoints. It comprises (1) a working memory that functions as a sliding window over recent frames to capture immediate context, and (2) an episodic memory that serves as a fixed-capacity bank for recall across frames. 
We design gated fusion and similarity-based update mechanisms to retain informative observations while preventing redundant accumulation, enabling coherent spatial reasoning across frames under bounded computation and storage.
We note that 3DLLM-Mem~\cite{hu20253dllmmem} augments VLMs with memory mechanisms. 
In contrast to memory architectures that grow stored representations over time, our dual-memory design integrates a sliding-window working memory for recent context with a fixed-capacity episodic memory that performs gated fusion.
This design ensures bounded memory usage while preserving reliable spatial reasoning. 

We evaluate \MethodName{} on multiple spatial reasoning and 3D understanding benchmarks, including VSI-Bench~\cite{yang2025thinking}, VSTI-Bench~\cite{fan2025vlm3r}, ScanQA~\cite{azuma2022scanqa}, and SQA3D \cite{ma2022sqa3d}.
\MethodName{} achieves superior performance among video-based models and surpasses open-source VLMs and spatial-enhanced models~\cite{bai2025qwen2,zhang2024llavavideo,wu2025spatialmllm,zheng2025vgllm,fan2025vlm3r}, advancing the frontier of visual-spatial intelligence.
Our main contributions include:
\begin{itemize}[topsep=0pt,itemsep=0pt,parsep=0pt]
    \item We introduce \MethodName{}, a vision-language model, integrating a semantic-geometric consistent representation with a persistent dual-memory, enabling spatial reasoning from video. \MethodName{} achieves state-of-the-art performance on multiple benchmarks, including VSI-Bench, VSTI-Bench, ScanQA, and SQA3D.
    \item We develop a 3D-aware representation for spatial reasoning that resolves semantic-geometric misalignment by grounding visual features into 3D space and enforcing cross-view consistency for coherent understanding. We also design a dual-memory module that couples a sliding-window working memory with a fixed-capacity episodic memory, achieving efficient spatial reasoning.
\end{itemize}

\section{Related Work}
\label{sec:related-work}

\subsection{3D Large Language Models}
Recent efforts have focused on enabling 3D large language models to understand 3D scenes by introducing explicit 3D modalities~\cite{chen2024ll3da,wang2023chat,huang2023embodied,huang2024chatscene,hong20233d,zhu2024llava,zheng2025video,wang2025ross3d,chen2025gsreasoner}.
Early work LL3DA~\cite{chen2024ll3da} employs a point cloud encoder for scene-level representations,
while Chat-3D~\cite{wang2023chat}, LEO~\cite{huang2023embodied}, and Chat-Scene~\cite{huang2024chatscene} segment objects and aggregate object-level 3D features for scenes.
3D-LLM~\cite{hong20233d} recovers 3D cues from rendered multi-view images via a 3D feature extractor, while LLaVA-3D~\cite{zhu2024llava} injects 3D position to form 3D-aware patch aggregation.
More recently, Video-3D LLM~\cite{zheng2025video} encodes 3D position into the visual representation.
Ross3D~\cite{wang2025ross3d} leverages reconstruction-based supervision to learn 3D-aware features.
GS-Reasoner~\cite{chen2025gsreasoner} constructs a 3D scene representation by integrating geometric features extracted from a point cloud encoder. 
These approaches demonstrate the effectiveness of incorporating explicit 3D information for 3D understanding.
In contrast, our framework focuses on learning 3D-aware representations from 2D inputs, requiring no additional 3D data or supervision.

\subsection{Spatial Reasoning in VLMs}
Equipping VLMs with spatial reasoning capabilities has recently emerged as a significant area of research~\cite{chen2024spatialvlm,cheng2024spatialrgpt,ma20253dsrbench,song2025robospatial,cai2025spatialbot,yang2025embodiedbench,yang2025magma}. However, previous work has mainly focused on spatial understanding from 2D static images, leaving video-based spatial reasoning as an underexplored domain.
To address this gap, VSI-Bench~\cite{yang2025thinking} introduces a video-based benchmark to evaluate how effectively VLMs can understand spatial relationships.
Recent studies~\cite{wu2025spatialmllm,zheng2025vgllm,fan2025vlm3r,huang2025mllms} have sought to enhance VLM spatial reasoning by incorporating geometric priors from  3D Visual Geometry Foundation Models (VGFMs).
For instance, Spatial-MLLM~\cite{wu2025spatialmllm} and VG-LLM~\cite{zheng2025vgllm} leverage VGFMs (e.g. VGGT~\cite{wang2025vggt}) as spatial encoders to extract geometric features. Similarly, 3DRS~\cite{huang2025mllms} introduces supervision signals from pretrained VGFMs, although this approach relies on additional 3D data to compute position information.
VLM-3R~\cite{fan2025vlm3r} incorporates implicit 3D tokens from pretrained VGFMs (e.g., CUT3R~\cite{wang2025cut3r}) and further proposes the VSTI-Bench to evaluate the comprehension of spatial relationships evolving over time.
Despite these advances, a common limitation of these methods is their tendency to simply fuse geometric features with semantic cues. Without explicitly addressing the potential semantic-geometric misalignment, their performance improvements remain constrained.
In contrast, our work introduces an explicit alignment mechanism to learn a view-consistent 3D-aware representation, which more effectively integrates semantic and geometric cues for spatial reasoning.

\subsection{Memory for Spatial Reasoning}
Memory mechanisms have been adopted in vision tasks that depend on long-range temporal or spatial information. Their applications include video understanding~\cite{he2024malmm,song2024moviechat,zhang2024flashvstream}, video generation~\cite{wu2025worldmemory,huang2025memoryforcing,yu2025context}, 3D reconstruction~\cite{wang2024span3r,wu2025point3r,chen2025long3r}, and robotic memory~\cite{wang2021unsupervised,wang2020visual,wang2024imperative,yang2025spatially}.
This paradigm has also been explored for lifelong navigation in embodied agents~\cite{khanna2024goat}, where memory representations are critical.
Within this embodied context, several works focus on building spatial memories.
For instance, MTU3D~\cite{zhu2025mtu3d} maintains a dynamic spatial memory bank for grounding and exploration, while 3D-Mem~\cite{yang20253dmem}
constructs a 3D scene memory from multi-view images for exploration and reasoning. More recently, 3DLLM-Mem~\cite{hu20253dllmmem} equips 3D LLMs with long-term memory to support diverse embodied tasks. However, this approach has limitations:
its unpruned memory accumulates redundant entries and incurs high computational cost on long videos.
In contrast, our work introduces a bounded dual-memory module to address these challenges. 
This module couples a sliding-window working memory with a fixed-capacity episodic memory, preserving salient information for efficient spatial reasoning.

\section{Method}
\label{sec:method}
\begin{figure*}[t]
\centering
\includegraphics[width=1.0\linewidth]{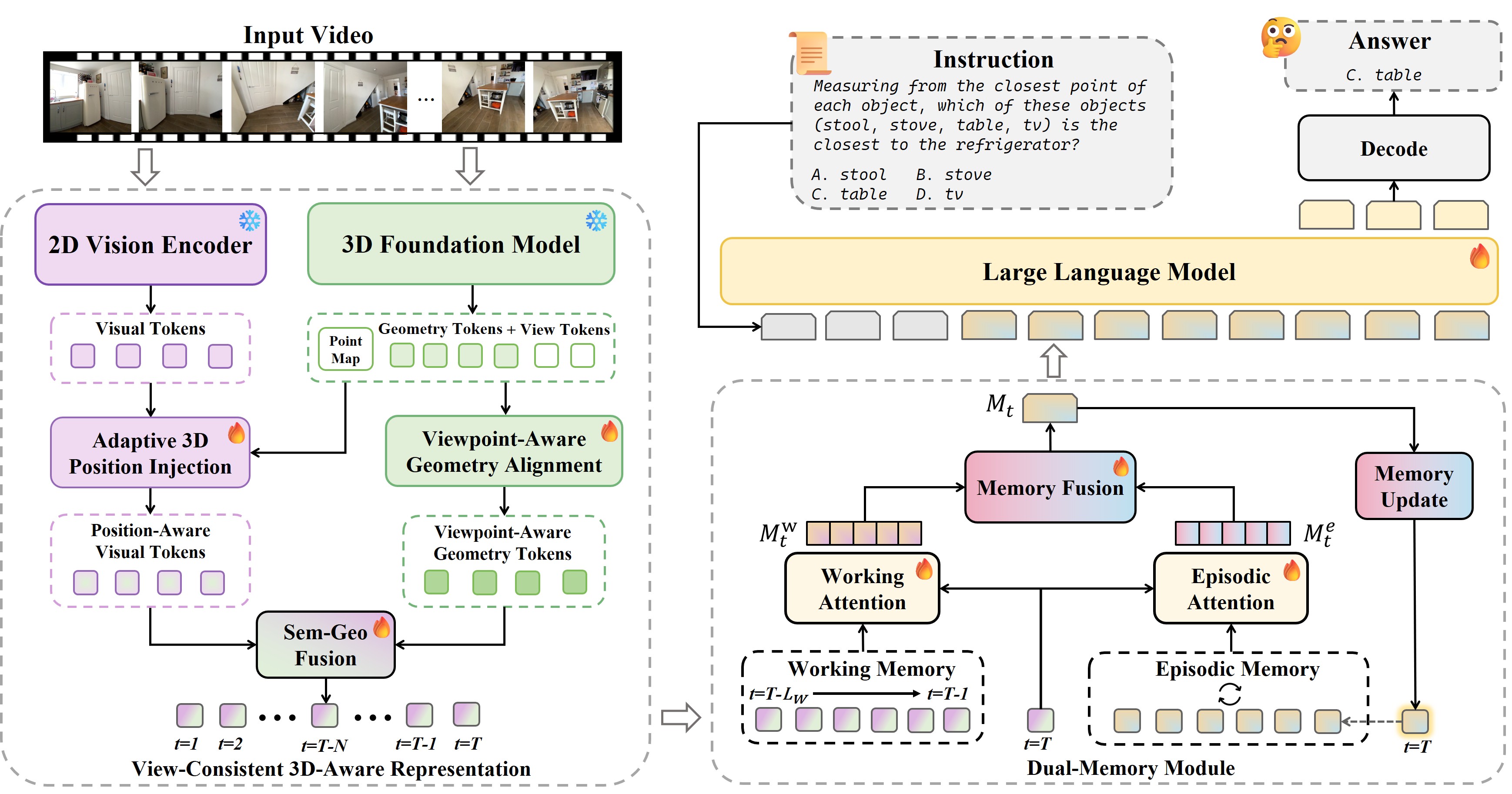}

\caption{\textbf{Overview of the \MethodName{} Architecture.} Our model constructs a view-consistent 3D-aware representation via adaptive 3D position injection, viewpoint-aware geometry alignment and semantic-geometric fusion. A dual-memory module with a sliding-window working memory and a fixed-capacity episodic memory maintains these representations over time, supporting temporally aware spatial reasoning.}
\label{fig:pipeline}
\end{figure*}

\noindent
\textbf{Overview}
We introduce \MethodName{}, a vision-language model that takes a monocular video $\mathcal{V}=\{{I}_t\}_{t=1}^{N}$ and a language instruction $Q$ as input, and generates answers by using a memory mechanism built directly upon view-consistent 3D-aware representations.
An overview of our architecture is shown in Fig.~\ref{fig:pipeline}.
Our method is built on two technical key innovations.
First, to address semantic-geometric misalignment, we explicitly align visual, geometry, and view tokens into a view-consistent 3D-aware representation (Sec.~\ref{sec:method-representation}) that maintains spatial consistency under camera motion.
Second, building on this consistent 3D-aware representation, we introduce a dual-memory module (Sec.~\ref{sec:method-memory}) that enables reasoning over both immediate and earlier observations.
This allows the model to reason about spatial layouts and objects no longer in view, supporting temporally aware and consistent spatial reasoning.
We implement \MethodName{} on top of LLaVA-Video~\cite{zhang2024llavavideo}. 
The 3D-aware, memory-enhanced representations are concatenated with the instruction embeddings and fed into the language backbone to generate answers.

\subsection{View-Consistent 3D-Aware Representation}
\label{sec:method-representation}
Given a video, the goal is to produce a view-consistent, globally coherent 3D-aware representation built from varying viewpoints, in which local semantic and geometric information is aligned for improved spatial reasoning. 
First, we uniformly sample $N$ frames $\{I_t\}_{t=1}^{N}$ from the input video, where $I_t\!\in\!\mathbb{R}^{H\times W\times 3}$.
From each frame $I_t$, we extract visual tokens $F_t\!\in\!\mathbb{R}^{h\times w\times c}$ capturing semantic information from a pretrained vision encoder, and geometric priors from a 3D foundation model ($\pi^{3}$~\cite{wang2025pi}): geometry tokens $G_t\!\in\!\mathbb{R}^{h\times w\times c_g}$, view tokens $Z_t \in \mathbb{R}^{h\times w\times c_v}$, and per-pixel point maps $X_t\!\in\!\mathbb{R}^{H\times W\times 3}$.
However, directly fusing these features suffers from a core semantic-geometric misalignment: the visual tokens $F_t$ lack spatial grounding, while the geometry tokens $G_t$ are viewpoint-ambiguous.
We introduce three core modules to address this issue: (1) \textit{Adaptive 3D Position Injection} injects predicted 3D coordinates into visual tokens while handling noisy predictions; (2) \textit{Viewpoint-Aware Geometry Alignment} aligns geometry tokens with their corresponding view tokens to resolve view ambiguity; and (3) \textit{Semantic-Geometric Fusion} fuses these aligned features into a 3D-aware representation that enables spatial reasoning consistent across camera motion.

\noindent
\textbf{Adaptive 3D Position Injection.}
Recent works such as Video-3D LLM~\cite{zheng2025video} enhance spatial awareness by injecting ground-truth 3D points (obtained from ground-truth depth maps and camera parameters) into the model.
This requirement limits their applicability in real-world scenarios, where reliable depth labels are often unavailable.
Moreover, we find that directly injecting predicted 3D coordinates (from a monocular depth estimation model) may introduce noise and inaccuracies, potentially affecting performance.
To address this, we introduce an adaptive gating mechanism that selectively incorporates reliable and useful 3D position cues (e.g. chairs instead of walls) while filtering out noisy or irrelevant predictions.
We first obtain per-patch 3D coordinates $C_t \in \mathbb{R}^{h\times w\times 3}$ by pooling the predicted point maps $X_t$ from a 3D foundation model, then encode them into 3D position embeddings $C_t' \in \mathbb{R}^{h \times w \times c}$ via a sinusoidal position encoding and a two-layer MLP $\phi:\mathbb{R}^{3}\!\rightarrow\!\mathbb{R}^{c}$.
The core of our adaptive design is a learnable gate $\alpha_t$ that modulates the injection:
\begin{equation}
    F_t^{\mathrm{pa}} = F_t + \alpha_t \odot \phi(C_t), 
    \qquad \alpha_t \in [0,1]^{h\times w\times 1},
\end{equation}
where $\odot$ denotes element-wise multiplication. This produces position-aware visual tokens $F_t^{\mathrm{pa}}$ grounded in 3D, while mitigating the impact of prediction noise.

\noindent
\textbf{Viewpoint-Aware Geometry Alignment.}
Direct fusing visual and geometric features presents a crucial challenge: the geometric features alone are viewpoint-ambiguous (e.g., a chair's front and back legs produce similar features). 
To make the geometric feature viewpoint-aware, we infuse geometry tokens with both patch-level and frame-level perspective cues.
First, to disambiguate local geometric patterns, we enrich geometry tokens with the corresponding view tokens, projected via a linear layer~$\psi_v:\mathbb{R}^{c_v}\!\to\!\mathbb{R}^{c_g}$
to produce $\hat{Z}_t$. This fusion produces viewpoint-aware features $G_t^{\mathrm{va}}$ that make patch-level geometry less ambiguous.
\begin{equation}
    G_t^{\mathrm{va}}=\mathrm{MLP}\big(\,\mathrm{Concat}[G_t\,;\,\hat{Z}_t]\,\big).
\end{equation}
Second, to provide global viewpoint context, we append a global view descriptor $\bar{Z}_t\in \mathbb{R}^{1\times 1\times c_v}$, obtained by pooling view tokens $Z_t$ and projecting via a linear layer $\psi_g:\mathbb{R}^{c_v}\!\to\!\mathbb{R}^{c_g}$. This provides a frame-level signal of the camera viewpoint.
\begin{equation}
    G_t^{\mathrm{vc}}=\mathrm{Concat}\big[\,G_t^{\mathrm{va}};\ Z_t^{g}\, \big]\
    \in \mathbb{R}^{(hw+1)\times c_g}.
\end{equation}
Infused with both patch-specific and frame-level viewpoint information, the resulting geometric tokens $G_t^{\mathrm{vc}}$ are now viewpoint-aware across different viewpoints.

\noindent
\textbf{Semantic-Geometric Fusion.}
To form the final 3D-aware representation for each frame $I_t$, we fuse the position-aware visual tokens and viewpoint-aware geometry tokens through cross-attention.
We use the position-aware visual tokens $F_t^{\mathrm{pa}}$ as queries and the viewpoint-aware geometry tokens $G_t^{\mathrm{vc}}$ as keys and values:
\begin{equation}
H_t = \mathrm{Attn}\big(Q(F_t^{\mathrm{pa}}), K(G_t^{\mathrm{vc}}), V(G_t^{\mathrm{vc}})\big).
\end{equation}
The output $H_t$ is a sequence of powerful 3D-aware tokens for frame $I_t$ that bind visual semantics to consistent geometric structure, resolving the semantic-geometric misalignment and producing a coherent 3D-aware representation that supports reliable spatial reasoning across views.

\subsection{Dual-Memory Module}
\label{sec:method-memory}
Inspired by CUT3R~\cite{wang2025cut3r}, which maintains a continuously updating state for 3D reconstruction, we propose a dual-memory module that stores and updates 3D-aware representations, enabling temporally spatial reasoning across frames. 

\noindent
\textbf{Working Memory for Immediate Retrieval.}
An agent's immediate environment is dense with information, 
but not all recent observations are equally relevant. 
To focus on important short-term context, we design a sliding-window working memory $\mathcal{W}_t$ to store the most recent $L_w$ representations. 
Our key insight is that the model should retrieve only what is relevant rather than treating all information equally. 
We achieve this selective retrieval by using the current representation $H_t$ as queries to attend over the working memory via cross-attention:
\begin{equation}
M_t^{w}=\mathrm{Attn}\big(Q(H_t),K(\mathcal{W}_t),V(\mathcal{W}_t)\big).
\end{equation}
The resulting feature $M_t^{w}$ is an enhanced representation of the current state, enriched with relevant context from the immediate past and recent observations.

\noindent
\textbf{Episodic Memory for Recall Across Frames.}
While the working memory captures recent context, it cannot retain crucial information beyond the immediate window, limiting spatial reasoning that depends on earlier observations.
For instance, the chair counting example in the introduction requires retrieving observations no longer stored in the sliding window.
To overcome this, we introduce an episodic memory $\mathcal{E}_t=\{E_1,E_2,...,E_{L_e}\}_{i=1}^{L_e}$, a fixed-capacity bank that stores the most salient observations across frames. Through end-to-end training on QA tasks, the model implicitly learns what constitutes a ``task-relevant'' memory, guiding it to preserve representations useful for spatial reasoning.
Similar to the working memory, we use the current representation $H_t$ to query episodic memory via cross-attention, yielding an episodic-enhanced representation $M_t^{e}$.

\noindent
\textbf{Memory Fusion and Update.}
The working-enhanced representation $M_t^w$ is fused with the episodic-enhanced representation $M_t^e$ to produce the final memory-enhanced feature for the current step.
We employ a learnable gate $\gamma_t$ to control the combination of these two information streams:
\begin{subequations}
\begin{align}
    &\gamma_t=\sigma\big(\mathrm{MLP}(\mathrm{Concat}[M_t^{w};M_t^{e}])\big),\\
    &M_t=\gamma_t \odot M_t^{w} + (1-\gamma_t) \odot M_t^{e}.
\end{align}
\end{subequations}
This produces the final memory-enhanced representation $M_t$.
To maintain a diverse and non-redundant episodic memory bank, we replace the most similar existing entry with current memory-enhanced representation $M_t$.
Specifically, we identify the most similar entry $E_{i_t^{\star}}$ in episodic memory and update it with $M_t$.
\begin{equation}
    i_t^{\star}=\mathrm{argmax}_{{i\in \{1,\ldots,L_e\}}} \mathrm{cos}(M_t,E_i).
\end{equation}
This similarity-based update mechanism ensures episodic memory remains bounded and diverse, supporting temporally aware spatial reasoning across frames.

\section{Experiments}
\noindent 
\textbf{Implementation Details.}
Our model is built on LLaVA-Video-7B~\cite{zhang2024llavavideo}, a video LLM based on Qwen2-7B~\cite{team2024qwen2}, and uses $\pi^3$~\cite{wang2025pi} as the 3D foundation model.
We train for one epoch on a mixed dataset using the same learning objective as VLM-3R~\cite{fan2025vlm3r}.
We use the AdamW optimizer with a batch size of 128 and a peak learning rate of 1e-5 for the LLM during the warmup phase.
For fine-tuning, we apply Low-Rank Adaptation (LoRA~\cite{hu2022lora}) with a rank of 128 and a scaling factor of 256. Both the vision encoder and 3D foundation model are kept fully frozen during training. 
All experiments are conducted on 8 NVIDIA H200 GPUs. 

\subsection{Spatial Reasoning Benchmarks}
\noindent 
\textbf{Datasets and Metrics.}
\begin{table*}[tb]
\centering
\caption{\textbf{Evaluations on VSI-Bench~\cite{yang2025thinking} for 3D spatial reasoning tasks.} We compare against proprietary models, open-sourced VLMs, and spatial-enhanced models specifically designed for spatial reasoning. Bold indicates the best performance.}
\resizebox{\linewidth}{!}
{
\begin{tabular}{l|cc|cccccccc}
\toprule
\multirow{2}{*}{\raisebox{-0.5ex}{\textbf{Methods}}}   &
\multirow{2}{*}{\raisebox{-0.5ex}{\textbf{Rank}}}      &
\multirow{2}{*}{\raisebox{-0.5ex}{\textbf{Avg.}}}      &
\multicolumn{4}{c}{\textbf{Numerical Qusetion}}        & 
\multicolumn{4}{c}{\textbf{Multiple-Choice Question}} \\
\cmidrule(lr){4-7}\cmidrule(lr){8-11}
& & & Obj. Cnt. & Abs. Dist. & Obj. Size & Room Size 
& Rel. Dist. & Rel. Dir. & Route Plan & Appr. Order \\
\midrule
\rowcolor{navyblue!5}
\multicolumn{11}{l}{\textcolor{black}{\textit{Proprietary Models (API)}}} \\
GPT-4o & 4 & 34.0 & 46.2 & 5.3 & 43.8 & 38.2 & 37.0 & 41.3 & 31.5 & 28.5 \\
GPT-5 & \cellcolor{oai-green-600}{1} & 55.0 & 53.3 & 34.5 & 73.3 & 47.5 & 63.7 & 48.7 & 50.3 & 68.9 \\
Gemini-2.5-Pro & \cellcolor{oai-green-200}{3} & 51.5 & 43.8 & 34.9 & 64.3 & 42.8 & 61.1 & 47.8 & 45.9 & 71.3 \\
Gemini-3-Pro & \cellcolor{oai-green-400}{2} & 52.5 & 38.0 & 37.8 & 72.7 & 44.1 & 59.9 & 55.7 & 45.9 & 66.0 \\
\midrule
\rowcolor{navyblue!5}
\multicolumn{11}{l}{\textcolor{black}{\textit{Open-sourced VLMs}}} \\
LongVA-7B & 13 & 29.2 & 38.0 & 16.6 & 38.9 & 22.2 & 33.1 & 43.3 & 25.4 & 15.7 \\
InternVL2-8B & 9 & 34.6 & 23.1 & {28.7} & 48.2 & {39.8} & 36.7 & 30.7 & 29.9 & 39.6 \\ 
InternVL2-40B & 7 & 36.0 & 34.9 & 26.9 & 46.5 & 31.8 & 42.1 & 32.2 & 34.0 & 39.6 \\
LongVILA-8B & 15 & 21.6 & 29.1 & 9.1 & 16.7 & 0.0 & 29.6 & 30.7 & 32.5 & 25.5 \\
VILA-1.5-8B & 14 & 28.9 & 17.4 & 21.8 & 50.3 & 18.8 & 32.1 & 34.8 & 31.0 & 24.8 \\
VILA-1.5-40B & 12 & 31.2 & 22.4 & 24.8 & 48.7 & 22.7 & 40.5 & 25.7 & 31.5 & 32.9 \\
Qwen2.5VL-7B & 10 & 33.0 & 40.9 & 14.8 & 43.4 & 10.7 & 38.6 & 38.5 & 33.0 & 29.8 \\
Qwen2.5VL-72B & 6 & 37.0 & 25.1 & 29.3 & 54.5 & 38.8 & 38.2 & 37.0 & 34.0 & 28.9 \\
Qwen3-VL-2B & \cellcolor{oai-green-200}{3} & 50.4 & 62.2 & 40.3 & 71.5 & 49.8 & 52.3 & 42.0 & 30.4 & 54.5 \\
Qwen3-VL-8B & \cellcolor{oai-green-600}{1} & 57.9 & 67.6 & 47.0 & 76.3 & 61.9 & 58.0 & 51.0 & 35.1 & 66.3 \\
Qwen3.5-4B & \cellcolor{oai-green-400}{2} & 53.6 & 56.5 & 36.5 & 67.5 & 53.8 & 60.3 & 57.5 & 34.0 & 62.3 \\
LLaVA-OneVision-7B & 11 & 32.4 & 47.7 & 20.2 & 47.4 & 12.3 & 42.5 & 35.2 & 29.4 & 24.4 \\
LLaVA-OneVision-72B & 5 & 40.2 & 43.5 & 23.9 & {57.6} & 37.5 & 42.5 & 39.9 & 32.5 & {44.6} \\
LLaVA-NeXT-Video-7B & 8 & 35.6 & 48.5 & 14.0 & 47.8 & 24.2 & {43.5} & 42.4 & 34.0 & 30.6 \\
LLaVA-NeXT-Video-72B & 4 & 40.9 & {48.9} & 22.8 & 57.4 & 35.3 & 42.4 & 36.7 & {35.0} & {48.6} \\
\midrule
\rowcolor{navyblue!5}
\multicolumn{11}{l}{\textcolor{black}{\textit{Spatial-Enhanced Models}}} \\
VG-LLM-4B & 5 & 47.3 & 66.0 & 37.8  & 55.2 & 59.2 & 44.6 & 45.6 & 33.5 & 36.4 \\
VG-LLM-8B & \cellcolor{oai-green-200}{3} & 50.7 & 67.9 & 37.7  & 58.6 & 62.0 & 46.6 & 40.7 & 32.4 & 59.2 \\
Spatial-MLLM-4B & 4 & 48.4 & 65.3 & 34.8  & 63.1 & 45.1 & 41.3 & 46.2 & 33.5 & 46.3 \\
VLM-3R-7B & \cellcolor{oai-green-400}{2} & 60.9 &{70.2} & {49.4} & {69.2} & {67.1} &{65.4} & {80.5} & {45.4} & 40.1 \\
\rowcolor{gray!10}
\textbf{VLM$^2$-7B (Ours)} & \cellcolor{oai-green-600}{1} & \textbf{68.8} & \textbf{72.5} & \textbf{59.6} & \textbf{70.8} & \textbf{69.9} & \textbf{69.0} & \textbf{87.8}& \textbf{52.6}&\textbf{68.3} \\
\midrule
\rowcolor{gray!10}
\textit{Improve} $\uparrow$ & - & \textcolor{ForestGreen}{+7.9}
& \textcolor{ForestGreen}{+2.3}
& \textcolor{ForestGreen}{+10.2}
& \textcolor{ForestGreen}{+1.6}
& \textcolor{ForestGreen}{+2.8}
& \textcolor{ForestGreen}{+3.6}
& \textcolor{ForestGreen}{+7.3}
& \textcolor{ForestGreen}{+7.2}
& \textcolor{ForestGreen}{+28.2}
\\
\bottomrule
\end{tabular}
}
\label{tab:vsibench}
\end{table*}
We evaluate our model's spatial reasoning performance on VSI-Bench~\cite{yang2025thinking}, which contains over 5,000 QA pairs collected from egocentric videos in ScanNet~\cite{dai2017scannet}, ScanNet++~\cite{yeshwanth2023scannet++}, and ARKitScenes~\cite{baruch2021arkitscenes}. 
We further report results on VSTI-Bench~\cite{fan2025vlm3r}, which is built on the same video sources and comprises approximately 6,000 QA pairs.
This benchmark evaluates the model's ability to interpret spatial layouts and reason about camera-object interactions under viewpoint changes in monocular videos.
For Multiple-Choice Answer (MCA) tasks, we report standard \textit{Accuracy}~\cite{metric1,metric2,metric3}, and for Numerical Answer (NC) tasks, we report \textit{Mean Relative Accuracy}~\cite{yang2025thinking}.

\noindent
\textbf{Baselines.}
We compare our model with a diverse set of proprietary and open-source VLMs (e.g., GPT-4o~\cite{hurst2024gpt4o}, Gemini-2.5 Pro~\cite{comanici2025gemini}, LLaVA-NeXT-Video~\cite{llavanext}, Qwen2.5-VL~\cite{bai2025qwen2}, Qwen3-VL~\cite{bai2025qwen3}). We also include recent spatial-enhanced models, including SPAR~\cite{zhang2025flatland}, VG-LLM~\cite{zheng2025vgllm}, Spatial-MLLM~\cite{wu2025spatialmllm}, and VLM-3R~\cite{fan2025vlm3r}.

\noindent
\textbf{Results on VSI-Bench.}
Table~\ref{tab:vsibench} presents the quantitative results on VSI-Bench. 
Our method consistently outperforms proprietary and open-source VLMs across all task categories.
The performance gains are particularly pronounced on \textit{Absolute Distance} and \textit{Relative Direction}, as these tasks demand strong spatial awareness and fine-grained 3D understanding.
This suggests that our model learns view-consistent and spatially-aware representations that effectively capture geometric cues.
Moreover, our method attains state-of-the-art results on \textit{Route Plan} and \textit{Appearance Order}. 
These tasks involve reasoning over changing viewpoints and maintaining consistent object relationships, highlighting the effectiveness of our representation and dual-memory module design.

\noindent
\textbf{Results on VSTI-Bench.}
As shown in Table~\ref{tab:vstibench}, we evaluate our model on VSTI-Bench, which assesses spatial reasoning under temporally evolving dynamics in monocular videos.
\begin{table}[tb]
\centering
\caption{\textbf{Evaluations on VSTI-Bench~\cite{fan2025vlm3r} for 3D spatial-temporal reasoning tasks. }
\textit{Abbr.:} CO-AbsD = Cam-Obj Abs. Dist.; CDisp = Cam. Displace.; CMDir = Cam. Mov. Dir.; OO-RelP = Obj-Obj Rel. Pos.; CO-RelD = Cam-Obj Rel. Dist. Bold indicates best performance in each model category.}
\resizebox{0.8\linewidth}{!}
{
\begin{tabular}{l|c|ccccc}
\toprule
\multirow{2}{*}{\raisebox{-0.5ex}{\textbf{Methods}}}   &
\multirow{2}{*}{\raisebox{-0.5ex}{\textbf{Avg.}}}      &
\multicolumn{2}{c}{\textbf{Numerical Question}}        & 
\multicolumn{3}{c}{\textbf{Multiple-Choice Question}} \\
\cmidrule(lr){3-4}\cmidrule(lr){5-7}
& & CO-AbsD & CDisp 
& CMDir & OO-RelP & CO-RelD \\
\midrule
\rowcolor{navyblue!5}
\multicolumn{7}{l}{\textcolor{black}{\textit{Proprietary Models (API)}}} \\
GPT-4o           & 38.2 & 29.5 & 23.4 & 37.3 & 58.1 & 42.5 \\ 
GPT-5 & 44.1 & 24.3 & 11.1 & 41.9 & 85.8 & 57.3 \\
Gemini-2.5-Pro & 42.4 & 29.2 & 8.0 & 30.0 & 84.0 & 60.7 \\
Gemini-3-Pro & 43.3 & 30.7 & 9.7 & 35.6 & 82.5 & 58.1 \\
\midrule
\rowcolor{navyblue!5}
\multicolumn{7}{l}{\textcolor{black}{\textit{Open-sourced VLMs}}} \\
LongVA-7B            & 32.3 & 13.5 & 5.1  & 43.7 & 57.9 & 41.2 \\
InternVL2-8B         & 43.5 & 32.9 & 13.5 & 48.0 & 68.0 & 55.0 \\ 
InternVL2-40B        & 43.2 & 11.9 & 34.9 & 33.3 & 63.8 & 72.2 \\ 
LongVILA-8B          & 30.5 & 20.0 & 11.6 & 35.4 & 52.3 & 33.4 \\
VILA-1.5-8B          & 37.3 & 30.1 & 27.3 & 42.2 & 50.4 & 36.7 \\
VILA-1.5-40B         & 38.2 & 28.2 & 15.7 & 28.8 & 65.4 & 53.0 \\
Qwen2.5VL-7B         & 38.2 & 22.9 & 4.9  & 47.4 & 65.9 & 49.9 \\
Qwen2.5VL-72B        & 40.3 & 18.0 & 10.0 & 41.0 & 74.2 & 58.4 \\
Qwen3-VL-2B & 45.4 & 30.2 & 35.0 & 37.9 & 65.0 & 59.0 \\
Qwen3-VL-8B & 47.0 & 26.1 & 31.8 & 37.1 & 79.7 & 60.1 \\
Qwen3.5-4B & 32.7& 26.0 & 24.8 & 24.9 & 71.9 & 16.0 \\
LLaVA-OneVision-7B   & 41.7 & 29.9 & 19.3 & 47.5 & 62.1 & 49.8 \\  
LLaVA-NeXT-Video-7B  & 40.0 & 28.2 & 1.8  & 49.8 & 64.7 & 55.6 \\
LLaVA-NeXT-Video-72B & 44.0 & 32.3 & 10.5 & 48.1 & 78.3 & 50.9 \\  
\midrule
\rowcolor{navyblue!5}
\multicolumn{7}{l}{\textcolor{black}{\textit{Spatial-Enhanced Models}}} \\
VLM-3R-7B & 58.8 & 39.4 & 39.6 & 60.6 & 86.5 & 68.6 \\
\rowcolor{gray!10}
\textbf{VLM$^2$-7B (Ours)} & \textbf{65.3} & \textbf{43.1} & \textbf{44.1} & \textbf{76.8} & \textbf{87.7} &\textbf{ 74.9} \\
\midrule
\rowcolor{gray!10}
\textit{Improve} $\uparrow$ 
& \textcolor{ForestGreen}{+6.5}
& \textcolor{ForestGreen}{+3.7}
& \textcolor{ForestGreen}{+4.5}
& \textcolor{ForestGreen}{+16.2}
& \textcolor{ForestGreen}{+1.2}
& \textcolor{ForestGreen}{+6.3}
\\
\bottomrule
\end{tabular}
}
\label{tab:vstibench}
\end{table}
Our method achieves state-of-the-art performance across all tasks, with an overall accuracy of 65.3, a relative improvement of 11.1\% over the previous best method VLM-3R (58.8), reflecting consistent gains on both spatial and temporal reasoning categories.
We attribute these gains to our view-consistent 3D-aware representation and dual-memory design, which support reasoning about camera motion and object interactions across frames.

\subsection{3D Scene Understanding Benchmarks}
\noindent
\textbf{Datasets and Metrics.}
We further evaluate our approach on 3D question answering benchmarks. We evaluate on ScanQA~\cite{azuma2022scanqa} for 3D spatial reasoning and SQA3D~\cite{ma2022sqa3d} for situated reasoning, both built on the ScanNet~\cite{dai2017scannet}. We follow the standard evaluation settings for each benchmark and use widely adopted evaluation metrics to assess answer quality in a consistent manner. For ScanQA, we report exact match accuracy (EM), CIDEr, BLEU-4, METEOR, and ROUGE-L. For SQA3D, we evaluate performance using exact match accuracy (EM).

\noindent
\textbf{Baselines.}
We compare against models covering different input settings and task specializations. 
Task-specific models including ScanQA~\cite{azuma2022scanqa}, SQA3D~\cite{ma2022sqa3d}, and 3D-VisTA~\cite{zhu20233d} are designed for 3D question answering tasks. 
We also compare with methods that operate on explicit 3D or 2.5D inputs, such as point clouds or depth maps, including Video-3D LLM~\cite{zheng2025video}, 3DRS~\cite{huang2025mllms}, and Ross3D~\cite{wang2025ross3d}. 
In addition, we compare with video-based models such as SPAR~\cite{zhang2025flatland} and VG-LLM~\cite{zheng2025vgllm}, which process multi-view video observations.

\noindent
\textbf{Results.}
We present the quantitative results on the ScanQA and SQA3D benchmarks in Table~\ref{tab:scanqasqa3d}. 
\begin{table}[tb]
\centering
\caption{\textbf{Evaluation on ScanQA~\cite{azuma2022scanqa} and SQA3D~\cite{ma2022sqa3d} for 3D understanding tasks.} ``C'' stands for ``CIDEr'', ``B-4'' for ``BLEU-4'', ``M'' for ``METEOR'', ``R'' for ``ROUGE'', and ``EM-1'' for top-1 exact match. \textbf{Bold} and \underline{underline} denote the best-performing and second-best performing models in each category, respectively.}
\setlength\tabcolsep{3pt} 
\resizebox{0.8\linewidth}{!}
{
\begin{tabular}{l|c|ccccc|c}
\toprule
\multirow{2}{*}{\raisebox{-0.5ex}{\textbf{Methods}}}
& \multirow{2}{*}{\makecell{\textbf{Video}\\\textbf{Input}}}
& \multicolumn{5}{c|}{\textbf{ScanQA (val)}}
& \textbf{SQA3D (test)}
\\
\cmidrule(lr){3-7} \cmidrule(lr){8-8}
& ~
& C & B-4 & M & R & EM-1 & EM-1 \\
\midrule 
\rowcolor{navyblue!5}
\multicolumn{8}{l}{\textcolor{black}{\textit{Task-Specific Models}}} \\
ScanQA~\cite{azuma2022scanqa} & \textcolor{Red}{\ding{55}} & \underline{64.9} & 10.1 & 13.1 & 33.3 & \underline{21.1} & \underline{47.2}  \\
SQA3D~\cite{ma2022sqa3d}                 & \textcolor{Red}{\ding{55}} & - & \textbf{11.2} & \underline{13.5} & \underline{34.5} & - & 46.6 \\
3D-VisTA~\cite{zhu20233d}     & \textcolor{Red}{\ding{55}} & \textbf{69.6} & \underline{10.4} & \textbf{13.9} & \textbf{35.7} & \textbf{22.4} & \textbf{48.5} \\
\midrule
\rowcolor{navyblue!5}
\multicolumn{8}{l}{\textcolor{black}{\textit{3D/2.5D-Input Models}}} \\
3D-LLM~\cite{hong20233d}         & \textcolor{Red}{\ding{55}} & 69.4 & 12.0 & 14.5 & 35.7 & 20.5 & -  \\
Chat-3D v2~\cite{huang2023chat}  & \textcolor{Red}{\ding{55}} & 87.6 & 14.0 & - & - & - & 54.7 \\
LL3DA~\cite{chen2024ll3da}       & \textcolor{Red}{\ding{55}} & 76.8 & 13.5 & 15.9 & 37.3 & -& -  \\
ChatScene~\cite{huang2024chatscene}   & \textcolor{Red}{\ding{55}} & 87.7 & 14.3 & 18.0 & 41.6 & 21.6 & 54.6 \\
LLaVA-3D~\cite{zhu2024llava}     & \textcolor{Red}{\ding{55}} & 103.1 & 16.4 & \underline{20.8} & 49.6 & \underline{30.6} & 60.1 \\
Video-3D LLM~\cite{zheng2025video}  & \textcolor{Red}{\ding{55}}   & 102.1 & 16.4 & 20.0 & 49.3 & 30.1 & 58.6\\
3DRS~\cite{huang2025mllms}       & \textcolor{Red}{\ding{55}} & \underline{104.8} & \underline{17.2} & 20.5 & \underline{49.8} & 30.3 & \underline{60.6} \\
Ross3D~\cite{wang2025ross3d}                   & \textcolor{Red}{\ding{55}} & \textbf{107.0} & \textbf{17.9} & \textbf{20.9} & \textbf{50.7} & \textbf{30.8} & \textbf{63.0} \\
\midrule 
\rowcolor{navyblue!5}
\multicolumn{8}{l}{\textcolor{black}{\textit{Video-Input Models}}} \\
InternVL2-8B~\cite{chen2024intervl8}   & \textcolor{Green}{\ding{51}} & 62.5 & 3.3 & 14.5 & 34.3 & - & 33.0 \\
Qwen2-VL-7B~\cite{bai2025qwen2}    & \textcolor{Green}{\ding{51}} & 53.9 & 3.0 & 11.4 & 29.3 & - & 46.5 \\
Qwen2-VL-72B~\cite{bai2025qwen2}    & \textcolor{Green}{\ding{51}} & 66.9 & 12.0 & 13.0 & 35.2 & - & 47.0 \\
LLaVA-Video-7B~\cite{zhang2024llavavideo} & \textcolor{Green}{\ding{51}} & 88.7 & 3.1 & 17.7 & 44.6 &  & 48.5\\
Oryx-34B~\cite{liu2024oryx}       & \textcolor{Green}{\ding{51}} & 72.3 & - & 15.0 & 37.3 & - & - \\
Spatial-MLLM-4B~\cite{wu2025spatialmllm} & \textcolor{Green}{\ding{51}} & \underline{91.8} & \underline{14.8} & \underline{18.4} & \underline{45.0} & - & \underline{55.9} \\
\rowcolor{gray!10}
\textbf{VLM$^2$-7B (Ours)}  & \textcolor{Green}{\ding{51}} & \textbf{105.5} & \textbf{17.7} & \textbf{20.5} & \textbf{50.3} & \textbf{30.7} & \textbf{60.4} \\
\midrule
\rowcolor{gray!10}
\textit{Improve} $\uparrow$ &
& \textcolor{ForestGreen}{+13.7}
& \textcolor{ForestGreen}{+2.9}
& \textcolor{ForestGreen}{+2.1}
& \textcolor{ForestGreen}{+5.3}
& \textcolor{ForestGreen}{-}
& \textcolor{ForestGreen}{+4.5}
\\
\bottomrule
\end{tabular}
}
\label{tab:scanqasqa3d}
\end{table}

Our method achieves state-of-the-art performance among video-based models on both benchmarks and also outperforms task-specific models. 
Compared with approaches that leverage explicit 3D or 2.5D inputs, Ross3D achieves better performance, as it exploits additional point clouds to render BEV images for reconstructive supervision. 
Our approach surpasses other 3D/2.5D input models, such as Video-3D LLM and 3DRS.
We attribute these gains to our model's ability to explicitly address semantic-geometric misalignment when constructing 3D-aware representations from video.
Our viewpoint-aware geometry alignment and adaptive 3D position injection jointly enforce alignment between visual and geometry tokens, leading to view-consistent 3D-aware representations.

\subsection{Analysis and Ablation Studies}
In this section, we provide further analysis of \MethodName{}, including its zero-shot generalization, performance across video durations, and detailed ablation studies.
We first evaluate it on additional benchmarks without fine-tuning, then analyze model performance across varying video durations, and finally examine the contribution of key components and memory configurations.

\noindent
\textbf{Zero-Shot Generalization.}
To evaluate zero-shot generalization, we assess \MethodName{} on three additional benchmarks: CV-Bench~\cite{tong2024cambrian}, SPAR-Bench~\cite{zhang2025flatland}, and BLINK~\cite{fu2024blink}.
Importantly, we do not perform any additional fine-tuning on these benchmarks.
\begin{table}[tb]
\centering
\caption{
\textbf{Zero-shot generalization on CV-Bench~\cite{tong2024cambrian}, SPAR-Bench~\cite{zhang2025flatland}, and BLINK~\cite{fu2024blink}. }We report accuracy without additional fine-tuning. 
}
\footnotesize
\setlength{\tabcolsep}{4pt}
\resizebox{0.9\linewidth}{!}{
\begin{tabular}{l|c|ccccccccc}
\toprule
\multirow{2}{*}{\raisebox{-0.5ex}{\textbf{Methods}}} &
\multirow{2}{*}{\raisebox{-0.5ex}{\textbf{Avg.}}} &
\multicolumn{3}{c}{\textbf{CV-Bench}} &
\multicolumn{4}{c}{\textbf{SPAR-Bench}} &
{\textbf{BLINK}}
\\
\cmidrule(lr){3-5}\cmidrule(lr){6-9} \cmidrule(lr){10-10} & &
2D & 3D & Avg. & Low & Medium & High & Avg. & Avg.\\
\midrule
GPT-4o~\cite{hurst2024gpt4o} & 57.3 & 69.4 & 81.3 & 75.4 & 29.3 & 24.9 & 45.1 & 36.4 & \underline{60.0} \\
Qwen2.5VL-3B~\cite{bai2025qwen2} & 47.3 & 69.1 & 72.2 & 70.6 & 15.3 & 26.4 & 32.2 & 24.6 & 46.6 \\
Qwen2.5VL-7B~\cite{bai2025qwen2} & 55.0 & \textbf{75.0} & 83.1 & 79.0 & 17.5 & 29.5 & 41.8 & 30.2 & 55.9 \\
VG-LLM-8B~\cite{zheng2025vgllm} & \underline{62.4} & 72.2 & \textbf{91.1} & \textbf{81.7} & \underline{39.6} & \underline{54.0} & \underline{71.7} & \underline{54.0} & 51.5 \\
\textbf{VLM$^2$-7B (Ours)} & \textbf{71.1} & \underline{73.6} & \underline{85.2} & \underline{79.4} & \textbf{58.6} & \textbf{71.2} & \textbf{78.8} & \textbf{66.8} & \textbf{67.2} \\
\bottomrule
\end{tabular}
}
\label{tab:zero_shot}
\end{table}

As shown in Table~\ref{tab:zero_shot}, \MethodName{} achieves competitive performance across all three benchmarks and attains the highest average results.
These results suggest that the proposed design learns transferable 3D-aware representations that generalize effectively beyond specific spatial reasoning benchmarks.

\noindent
\textbf{Performance Across Video Durations.}
To analyze how performance varies with video duration, we partition the input videos in VSI-Bench~\cite{yang2025thinking} and VSTI-Bench~\cite{fan2025vlm3r} into three groups based on length: Short ($<1$ min), Mid (1-2 min), and Long ($>2$ min).
\begin{table}[tb]
\centering
\caption{
\textbf{Spatial reasoning performance on VSI-Bench~\cite{yang2025thinking} and VSTI-Bench~\cite{fan2025vlm3r} across varying video durations. }Videos are grouped into Short ($<1$ min), Mid (1--2 min), and Long ($>2$ min). 
}
\footnotesize
\setlength{\tabcolsep}{4pt}
\resizebox{0.85\linewidth}{!}{
\begin{tabular}{l|cccc|cccc}
\toprule
\multirow{2}{*}{\raisebox{-0.5ex}{\textbf{Methods}}} &
\multicolumn{4}{c|}{\textbf{VSI-Bench}} &
\multicolumn{4}{c}{\textbf{VSTI-Bench}} \\
\cmidrule(lr){2-5}\cmidrule(lr){6-9} &
Avg. & Short & Mid & Long & Avg. & Short & Mid & Long \\
\midrule
\rowcolor{navyblue!5}
\multicolumn{9}{l}{\textcolor{black}{\textit{Open-sourced VLMs}}} \\
Qwen2.5VL-7B~\cite{bai2025qwen2} & 29.5 & 34.3 & 28.3 & 28.4 & 45.9 & 43.3 & 48.4 & 45.5 \\
Qwen2.5VL-72B~\cite{bai2025qwen2} & 36.3 & 39.2 & 36.6 & 34.7 & 49.4 & 45.1 & 50.8 & 55.2 \\
LLaVA-NeXT-Video-7B~\cite{llavanext} & 36.8 & 38.5 & 37.3 & 35.5 & 46.4 & 43.0 & 47.9 & 50.1 \\
LLaVA-NeXT-Video-72B~\cite{llavanext} & 41.1 & 41.3 & 40.7 & 41.4 & 51.3 & 47.6 & 51.9 & 57.6 \\
\midrule
\rowcolor{navyblue!5}
\multicolumn{9}{l}{\textcolor{black}{\textit{Spatial-Enhanced Models}}} \\
Spatial-MLLM-4B~\cite{wu2025spatialmllm} & 48.1 & 49.8 & 48.3 & 47.2 & - & - & - & - \\
VG-LLM-4B~\cite{zheng2025vgllm} & 47.4 & 49.1 & 48.7 & 45.2 & - & - & - & - \\
VG-LLM-8B~\cite{zheng2025vgllm} & 50.5 & 51.9 & 49.7 & 50.6 & - & - & - & - \\
VLM-3R-7B~\cite{fan2025vlm3r} & \underline{63.2} & \underline{67.1} & \underline{64.7} & \underline{60.0} & \underline{63.6} & \underline{61.1} & \underline{64.7} & \underline{66.4} \\
\textbf{VLM$^2$-7B (Ours)} & \textbf{69.4} & \textbf{71.0} & \textbf{70.0} & \textbf{68.1} & \textbf{69.5} & \textbf{65.9} & \textbf{70.6} & \textbf{74.6} \\
\midrule
\rowcolor{gray!10}
\textit{Improve} $\uparrow$
& \textcolor{ForestGreen}{+6.2}
& \cellcolor{oai-green-200}{+3.9}
& \cellcolor{oai-green-400}{+5.3}
& \cellcolor{oai-green-600}{+8.1}
& \textcolor{ForestGreen}{+5.9}
& \cellcolor{oai-green-200}{+4.8}
& \cellcolor{oai-green-400}{+5.9}
& \cellcolor{oai-green-600}{+8.2}
\\
\bottomrule
\end{tabular}
}
\label{tab:long_horizon}
\end{table}

Table~\ref{tab:long_horizon} shows the comparisons. 
On VSI-Bench, VLM-3R's performance degrades as video length increases, 
indicating challenges in maintaining consistent spatial reasoning under longer sequences, whereas our model shows more stable performance across durations.
On VSTI-Bench, while VLM-3R's performance remains relatively stable across lengths, our model achieves higher accuracy in all groups, with larger gains on long videos.
Overall, VLM$^2$ improves average accuracy by $+6.2$ and $+5.9$ points on VSI-Bench and VSTI-Bench, respectively, with the larger gains on long videos.
These results suggest that our view-consistent 3D-aware representation and dual-memory design contribute to reliable spatial reasoning across varying video durations.

\noindent
\textbf{Overall Component Analysis.}
Our component ablation studies are summarized in Table~\ref{tab:ablation_overall}.
\begin{table}[tb]
\centering
\caption{
\textbf{Ablation of model components.}
We evaluate the effect of 3D-aware representation, working memory, and episodic memory.
Each module contributes complementary improvements, and combining all yields the best overall performance.
}
\small
\setlength\tabcolsep{3pt} 
\resizebox{\columnwidth}{!}{
\begin{tabular}{ccc|c|ccccc}
\toprule
\multirow{2}{*}{\makecell[c]{\textbf{3D-Aware}\\\textbf{Rep.}}} &
\multirow{2}{*}{\makecell[c]{\textbf{Work.}\\\textbf{Mem.}}} &
\multirow{2}{*}{\makecell[c]{\textbf{Epis.}\\\textbf{Mem.}}} &
\multirow{2}{*}{\raisebox{-0.5ex}{\textbf{Avg.}}} &
\multicolumn{2}{c}{\textbf{Numerical Question}} &
\multicolumn{3}{c}{\textbf{Multiple-Choice Question}} \\
\cmidrule(lr){5-6}\cmidrule(lr){7-9} &
~ & ~ &  ~ & Abs. Dist. & Room Size & Rel. Dist. & Rel. Dir. & Route Plan\\
\midrule
\ding{55} & \ding{55} & \ding{55} & 55.2 & 43.3 & 62.4 & 62.1 & 67.8 & 40.2 \\
\ding{51} & \ding{55} & \ding{55} & 63.8 & 52.9 & 67.5 & 65.6 & 84.8 & 48.5 \\
\ding{51} & \ding{51} & \ding{55} & 65.9 & 56.3 & 68.3 & 67.3 & 86.2 & 51.3 \\
\ding{51} & \ding{55} & \ding{51} & 66.1 & 57.7 & 68.6 & 67.6 & 85.9 & 50.5 \\
\ding{51} & \ding{51} & \ding{51} & \textbf{67.8} & \textbf{59.6} & \textbf{69.9} & \textbf{69.0} & \textbf{87.8 }& \textbf{52.6} \\
\bottomrule
\end{tabular}
}
\label{tab:ablation_overall}
\end{table}
We establish a strong baseline by fine-tuning LLaVA-NeXT-Video-7B~\cite{llavanext} with the same language backbone, training schedule, and compute budget for all variants.
Incorporating our 3D-aware representation into the baseline yields an average of 8.6\% accuracy gain.
Building on this representation,
introducing working memory or episodic memory brings additional gains of 2.1\% and 2.3\%, respectively.
When both memory modules are combined, our full model achieves the best performance, improving by 4.0\% over the 3D-aware representation and by 12.6\% over the baseline.
The benefit is particularly pronounced on tasks such as \textit{Route Plan} (+ 12.4), which requires spatial structure awareness, highlighting the importance of an explicit memory mechanism.
These findings suggest that the view-consistent 3D-aware representation and the dual-memory module are crucial components for advanced spatial reasoning.

\noindent
\textbf{Effectiveness of 3D-Aware Representation.}
\begin{table}[tb]
\centering
\caption{\textbf{Ablation of 3D-aware representations.} We compare different 3D foundation models (VGGT~\cite{wang2025vggt}, CUT3R~\cite{wang2025cut3r}, $\pi^3$~\cite{wang2025pi}) and fusion strategies (\textit{Concat-MLP}, \textit{Cross-Attn}). Built on $\pi^{3}$ and \textit{Cross-Attn}, our viewpoint-aware geometry alignment (VAGA) and adaptive 3D position injection (A3PI) achieve the best results.
}
\small
\setlength{\tabcolsep}{5pt}
\renewcommand{\arraystretch}{1.1}
\setlength\tabcolsep{3pt}
\resizebox{\columnwidth}{!}
{
\begin{tabular}{l|c|ccccc}
\toprule
\multirow{2}{*}{\raisebox{-0.5ex}{\textbf{Methods}}}  &
\multirow{2}{*}{\raisebox{-0.5ex}{\textbf{Avg.}}} &
\multicolumn{2}{c}{\textbf{Numerical Question}} &
\multicolumn{3}{c}{\textbf{Multiple-Choice Question}} \\
\cmidrule(lr){3-4}\cmidrule(lr){5-7} &
~ & Abs. Dist. & Room Size & Rel. Dist. & Rel. Dir. & Route Plan\\
\midrule
\rowcolor{navyblue!5}
\multicolumn{7}{l}{\textcolor{black}{\textit{Baseline}}} \\
LLaVA-NeXT-Video-SFT & 55.2 & 43.3 & 62.4 & 62.1 & 67.8 & 40.2 \\
\midrule
\rowcolor{navyblue!5}
\multicolumn{7}{l}{\textcolor{black}{\textit{Sem-Geo Fusion (Concat-MLP)}}} \\
Baseline + VGGT  & 41.4 & 38.6 & 45.6 & 38.7 & 47.6 & 36.6 \\
Baseline + CUT3R  & 43.3 & 39.8& 51.4 & 41.3 & 45.9 & 38.1 \\
Baseline + $\pi^3$  & 40.4 & 38.3 & 42.5 & 41.0 & 45.2 & 35.1\\
\midrule
\rowcolor{navyblue!5}
\multicolumn{7}{l}{\textcolor{black}{\textit{Sem-Geo Fusion (Cross-Attn)}}} \\
Baseline + VGGT   & 60.2 & 51.0 & 63.9 & 61.4 & 81.8 & 42.7 \\
Baseline + CUT3R  & 59.4 & 50.2 & 63.1 & 62.8 & 79.3 & 41.8 \\
Baseline + $\pi^3$ & 61.0 & 51.9 &  64.7 & 63.1 & 81.9 & 43.3 \\
\midrule
\rowcolor{navyblue!5}
\multicolumn{7}{l}{\textcolor{black}{\textit{VLM$^2$ (Ours) }{\scriptsize [3D Foundation: $\pi^{3}$; Sem-Geo Fusion: Cross-Attn]}}}\\

Baseline + A3PI (w/o adapt.)  & 58.9 & 50.3 & 64.5 & 60.5 & 77.2 & 41.9 \\
Baseline + A3PI  & 61.6 & 52.3 & 65.3 & 63.7 & 82.1 & 44.7  \\
Baseline + VAGA  & 62.9 & 52.5 & 66.4 & 64.9 & 84.0 & 46.6 \\
Baseline + A3PI + VAGA  & 63.8 & 52.9 & 67.5 & 65.6 & 84.8 & 48.5 \\
\bottomrule
\end{tabular}
}
\label{tab:ablation_overall_signle}
\end{table}

We ablate the choice of 3D backbone and fusion strategy on top of the aforementioned baseline, as shown in Table~\ref{tab:ablation_overall_signle}.
To incorporate geometry tokens from 3D foundation models into visual tokens, we compare two fusion strategies.
A simple \textit{Concat-MLP} fusion, which concatenates 2D visual tokens with 3D geometry tokens followed by an MLP, yields lower performance (40.4--43.3), indicating that naïve concatenation leaves semantic-geometric misalignment unresolved.
Replacing Concat-MLP with a \textit{Cross-Attn} fusion using 3D backbones (VGGT, CUT3R, $\pi^3$) raises performance to 59.4--61.0, showing that geometric cues help with a stronger fusion strategy but still fall short of our approach.
Our viewpoint-aware geometry alignment (VAGA) with adaptive 3D position injection (A3PI) achieves the best accuracy, boosting performance from 61.0 to 63.8.
Removing the adaptive mechanism from A3PI reduces the performance to 58.9, indicating that injecting position information into all visual tokens introduces noise in irrelevant regions and distorts feature distributions, highlighting the importance of adaptive injection.

\noindent
\textbf{Ablation of Dual-Memory Length.}
We study the effect of memory length by varying the working memory window $L_w$ and episodic memory capacity $L_e$. 
\begin{table}[tb]
\centering
\footnotesize 
\caption{
\textbf{Ablation of dual-memory length.}
We vary the length of working memory ($L_w$) and episodic memory ($L_e$) on VSI-Bench to study the effect of memory capacity. The setting $(L_w, L_e){=}(8,32)$ achieves the best overall performance.
}
\setlength{\tabcolsep}{4pt}
\resizebox{0.6\linewidth}{!}{
\begin{tabular}{cc|ccc}
\toprule
\multicolumn{2}{c|}{\textbf{Memory Size}} & 
\multicolumn{3}{c}{\textbf{VSI-Bench}} \\
\cmidrule(lr){1-2}\cmidrule(lr){3-5}
$L_w$ & $L_e$ & Numerical & Multiple-Choice & Avg. \\
\midrule
4 & 8 & 64.1 & 58.5 & 61.3 \\
8 & 8 & 65.6 & 66.9 & 66.3\\
8 & 16 & 67.1 & 67.4 & 67.3\\
8 & 32 & \textbf{68.2} & \textbf{69.4 }& \textbf{68.8}\\
16 & 8 & 65.2 & 65.7 & 65.5\\
16 & 16 & 67.3 & 67.9 & 67.6\\
16 & 32 & 67.9 & 68.6 & 68.3\\
\bottomrule
\end{tabular}
}
\label{tab:ablation_memory_length}
\end{table}

As reported in Table~\ref{tab:ablation_memory_length}, increasing episodic memory capacity $L_e$ improves performance up to $L_e=32$, highlighting the role of episodic capacity. 
In contrast, a larger working memory window  $L_w$ is not always helpful. 
Working memory attends to recent frames, so a short window may miss spatial cues needed to disambiguate nearby objects, whereas a long window introduces irrelevant tokens that dilute attention. 
We find that $L_w=8$, $L_e=32$ achieves the best performance, 
indicating an effective balance between the two memory components.

\section{Conclusion}
In this paper, we propose VLM$^2$, a vision-language model for video-based spatial reasoning that addresses two critical challenges: semantic-geometric misalignment and the absence of persistent memory.
VLM$^2$ constructs a view-consistent 3D-aware representation from video by adaptively grounding visual features into 3D space and enforcing cross-view consistency, enabling coherent 3D understanding.
On top of this representation, we introduce a dual-memory module that combines a sliding-window working memory for immediate context with a fixed-capacity episodic memory for recall across frames, allowing efficient temporally aware spatial reasoning.
Extensive experiments across multiple spatial reasoning benchmarks demonstrate that our method achieves state-of-the-art performance, advancing the frontier of visual-spatial intelligence.

\bibliographystyle{splncs04}
\bibliography{main}
\clearpage
\appendix

\begin{center}
{\Large \textbf{Vision-Language Memory for Spatial Reasoning}}\\
\vspace{1.0em}
{\Large Supplementary Material}
\end{center}
\vspace{1em}

\definecolor{darkgreen}{rgb}{0.0, 0.5, 0.0}
\definecolor{blueviolet}{rgb}{0.54, 0.17, 0.89}
\definecolor{DarkBlue}{rgb}{0.0, 0.0, 0.55}

\vspace{-2.5em}
\section{Details of Dual-Memory Module}
\vspace{-2.5em}
\definecolor{global}{RGB}{21,96,130}
\definecolor{breakpoint}{RGB}{51,0,111}
\algnewcommand{\Comment}[1]{\textcolor{blue}{\(\triangleright\) #1}}

\begin{algorithm}[h]
\caption{Dual-Memory Module}\label{alg:memory}
\begin{algorithmic}[1]
\Require 3D-aware representation $H_t$, working memory $\mathcal{W}_t$, episodic memory $\mathcal{E}_t$

\Ensure Memory-enhanced representation $M_t$, updated $\mathcal{W}_{t+1}$, updated $\mathcal{E}_{t+1}$

\State \textcolor{global}{\(\triangleright\) Working Retrieval}
\State $M_t^{w}\gets \mathrm{Working~Attention}({Q}=H_t,{KV}=\mathcal{W}_t)$

\State \textcolor{global}{\(\triangleright\) Episodic Retrieval}
\State $M_t^{e}\gets \mathrm{Episodic~Attention}({Q}=H_t,{KV}=\mathcal{E}_t)$

\State \textcolor{global}{\(\triangleright\) Gated Memory Fusion}
\State $\gamma_t=\sigma(\mathrm{MLP}(\mathrm{Concat}[M_t^{w};M_t^{e}]))$ 
\State $M_t=\gamma_t\odot M_t^{w}+(1-\gamma_t)\odot M_t^{e}$

\State \textcolor{global}{\(\triangleright\) Update Working Memory} 
\If{$|\mathcal{W}_t|<L_w$}
    \State $\mathcal{W}_{t+1}\gets \mathcal{W}_t \cup \{H_t\}$
\Else
    \State \textbf{remove} oldest element from $\mathcal{W}_t$
    \State $\mathcal{W}_{t+1}\gets \mathcal{W}_t \cup \{H_t\}$
\EndIf

\State \textcolor{global}{\(\triangleright\) Update Episodic Memory} 
\If{$|\mathcal{E}_t|<L_e$}
    \State $\mathcal{E}_{t+1}\gets \mathcal{E}_t \cup \{M_t\}$
\Else
    \For{$i=1$ \textbf{to} $L_e$} 
        \State $s_i \gets \mathrm{cos}(M_t, E_i)$
    \EndFor
    \State $i_t^{\star}\gets \arg\max_{i\in\{1,\dots,L_e\}} s_i$
    \State \textbf{del} $E_{i_t^{\star}}$;~~$\mathcal{E}_{t+1}\gets \mathcal{E}_t \cup \{M_t\}$
\EndIf

\State \Return $M_t,~\mathcal{W}_{t+1},~\mathcal{E}_{t+1}$
\end{algorithmic}
\end{algorithm}
\vspace{-2.0em}

\section{Implementation Details}

\subsection{Training Datasets}
Our VLM$^2$ is a spatial reasoning model capable of solving multiple spatial tasks.
To achieve this capability, we construct a meticulously curated mixed dataset that combines spatial reasoning, spatial-temporal reasoning, and 3D scene understanding question-answering (QA) pairs.

\noindent
\textbf{Spatial Reasoning QA.}
We first include spatial reasoning datasets from VLM-3R~\cite{fan2025vlm3r}, which focus on visual-spatial intelligence from egocentric videos. These QA pairs cover object count, absolute distance, object size, room size, relative distance, relative direction, and appearance order.

\noindent
\textbf{Spatial-Temporal Reasoning QA.}
To advance spatial-temporal reasoning in 3D environments, we further incorporate the spatial-temporal QA pairs from VLM-3R~\cite{fan2025vlm3r}. These questions interrogate camera dynamics, object states, and complex camera-object interactions over time, encompassing camera displacement, camera-object absolute distance, camera-object relative distance, object-object relative position, and camera movement direction.

\noindent
\textbf{3D Scene Understanding QA.}
We also include 3D scene understanding datasets from ScanQA~\cite{azuma2022scanqa} and SQA3D~\cite{ma2022sqa3d}. 
ScanQA provides 23K QA pairs about object alignment, directions, and object localization. 
SQA3D contributes approximately 79K situated QA pairs, where an agent in a 3D scene needs to infer and localize its situation (position, orientation, \textit{etc}.) from textual descriptions, and then answer questions that require strong situated spatial reasoning.
Together, these datasets enable VLM$^2$ to effectively learn both video-based spatial reasoning and 3D scene understanding capabilities.

\subsection{Training Details}
Our model is built on LLaVA-Video-7B~\cite{zhang2024llavavideo}, initialized from the pretrained checkpoint \textit{LLaVA-Video-7B-Qwen2}. 
The vision encoder is initialized with \textit{siglip-so400m-patch14-384}, and we adopt $\pi^3$~\cite{wang2025pi} as the 3D foundation model. 
During training, both the vision encoder and the 3D foundation model are kept frozen, while the language backbone and our modules are updated.
We set the gradient accumulation steps to 8, use the AdamW optimizer with a global batch size of 128 and a peak learning rate of 1e-5 for the LLM during the warmup phase, and uniformly sample 32 frames per scene for video-based input. 
For fine-tuning, we apply Low-Rank Adaptation (LoRA~\cite{hu2022lora}) with a rank of 128 and a scaling factor of 256, and employ DeepSpeed ZeRO-2~\cite{rajbhandari2020zero} for memory optimization.
All experiments are conducted on 8 NVIDIA H200 GPUs.

\subsection{Evaluation Details}
To assess the spatial reasoning capability of VLM$^2$ across a diverse set of tasks, we evaluate it on four widely adopted benchmarks.
We conduct all evaluations using the LLMs-Eval~\cite{zhang2025lmmseval} project.
For video-based benchmarks, we uniformly sample 32 frames per scene from the input video.

\noindent
\textbf{VSI-Bench.}
We use VSI-Bench~\cite{yang2025thinking} to evaluate the model's spatial reasoning performance from egocentric videos.
We follow the official evaluation protocol of VSI-Bench, adopting a greedy decoding strategy for all models to ensure fair comparison.
Moreover, we keep the question templates, prompt formats, and task-specific instructions exactly the same as in the original benchmark.

\noindent
\textbf{VSTI-Bench.}
We also evaluate our model on VSTI-Bench~\cite{fan2025vlm3r}, which is built on the same video sources and contains approximately 6,000 QA pairs.
This benchmark evaluates the model's ability to interpret spatial layouts and reason about camera-object interactions under viewpoint changes in monocular videos.
We follow the same evaluation settings as VLM-3R~\cite{fan2025vlm3r}, including task definitions, prompt formats, and decoding configurations.

\noindent
\textbf{ScanQA and SQA3D.}
For 3D scene understanding, we also evaluate our model on ScanQA~\cite{azuma2022scanqa} and SQA3D~\cite{ma2022sqa3d}.
During inference, we set the number of frames to 32, following the evaluation configuration of Video-3D LLM~\cite{zheng2025video}, so as to maintain a comparable video context across methods.

\section{Additional Experimental Results}
\subsection{Additional Results on ScanQA and SQA3D}
\begin{table*}[!t]
\caption{\textbf{Additional evaluation results on ScanQA~\cite{azuma2022scanqa} for 3D understanding tasks.} \textbf{Bold} and \underline{underline} denote the best-performing and second-best performing models in each category, respectively.}
\centering
\resizebox{\textwidth}{!}
{
    \begin{tabular}{l|c|cccccccc}
    \toprule
    \multirow{2}{*}{\raisebox{-0.5ex}{\textbf{Methods}}}
    & \multirow{2}{*}{\makecell{\textbf{Video}\\\textbf{Input}}}
    & \multicolumn{8}{c}{\textbf{ScanQA (val)}}
    \\
    \cmidrule(lr){3-10}
    & ~
    & EM-1 & BLEU-1 & BLEU-2  & BLEU-3 & BLEU-4 & ROUGE-L & METEOR & CIDEr\\
    \midrule 
    \rowcolor{navyblue!5}
    \multicolumn{10}{l}{\textcolor{black}{\textit{Task-Specific Models}}} \\
    ScanQA~\cite{azuma2022scanqa}     & \textcolor{Red}{\ding{55}} & \underline{21.1} & 30.2 & 20.4 & 15.1 & \underline{10.1} & \underline{33.3} & \underline{13.1} & \underline{64.9} \\
    3D-VisTA~\cite{zhu20233d}   & \textcolor{Red}{\ding{55}}  & \textbf{22.4} & - & - & - & \textbf{10.4} & \textbf{35.7} & \textbf{13.9} & \textbf{69.6} \\
    \midrule
    \rowcolor{navyblue!5}
    \multicolumn{10}{l}{\textcolor{black}{\textit{3D/2.5D-Input Models}}} \\
    ChatScene~\cite{huang2024chatscene} & \textcolor{Red}{\ding{55}} & {21.6} & {43.2} & {29.1} & {20.6} & 14.3 & 41.6 & 18.0 & 87.7 \\
    LLaVA-3D~\cite{zhu2024llava}        & \textcolor{Red}{\ding{55}} & 27.0 & - & - & - & 14.5 & \underline{50.1} & \underline{20.7} & 91.7 \\
    Video-3D LLM~\cite{zheng2025video}  & \textcolor{Red}{\ding{55}} & {30.1} & {47.1} & {31.7} & {22.8} &{16.2} & {49.0} & {19.8} & {102.1} \\
    3DRS~\cite{huang2025mllms}        & \textcolor{Red}{\ding{55}} & \underline{30.3} & \underline{48.4} & \underline{32.7} & \underline{23.8} & \underline{17.2} & 49.8 & 20.5 & \underline{104.8}\\
    Ross3D~\cite{wang2025ross3d}       & \textcolor{Red}{\ding{55}} & \textbf{30.8} & \textbf{49.2} & \textbf{33.7} &\textbf{ 24.9} & \textbf{17.9} &\textbf{ 50.7} & \textbf{20.9} & \textbf{107.0} \\
    \midrule 
    \rowcolor{navyblue!5}
    \multicolumn{10}{l}{\textcolor{black}{\textit{Video-Input Models}}} \\
    Qwen2-VL-7B~\cite{bai2025qwen2}          & \textcolor{Green}{\ding{51}}  & 19.0 & 27.8 & 13.6 & 6.3 &  3.0 & 29.3 & 11.4 & 53.9 \\
    Qwen2-VL-72B~\cite{bai2025qwen2}        & \textcolor{Green}{\ding{51}} & 24.0 & 26.8 & 17.8 & 14.6 &  12.0 & 35.2 & 13.0 & 66.9 \\
    LLaVA-Video-7B~\cite{zhang2024llavavideo}  & \textcolor{Green}{\ding{51}} & - & 39.7 & 26.6 & 9.3 & 3.1 & {44.6} & 17.7 & 88.7 \\
    Oryx-34B~\cite{liu2024oryx}  & \textcolor{Green}{\ding{51}} & - & 38.0 & 24.6 & - & - & 37.3 & 15.0 & 72.3\\
    Spatial-MLLM-4B~\cite{wu2025spatialmllm}  & \textcolor{Green}{\ding{51}} & \underline{26.3} & \underline{44.4} & \underline{28.8} & \underline{21.9} & \underline{14.8} & \underline{45.0} &\underline {18.4} & \underline{91.8} \\
    \rowcolor{gray!10}
    \textbf{VLM$^2$-7B (Ours)}     & \textcolor{Green}{\ding{51}} & \textbf{30.7} & \textbf{48.7} & \textbf{33.1} & \textbf{24.5} & \textbf{17.7} & \textbf{50.3} & \textbf{20.5} & \textbf{105.5}\\
    \bottomrule
    \end{tabular}
}
\label{tab:fullscanqa}
\vspace{-1.5em}
\end{table*}

\begin{table*}[!t]
\caption{\textbf{Additional evaluation results on SQA3D~\cite{ma2022sqa3d} for 3D understanding tasks.} \textbf{Bold} and \underline{underline} denote the best-performing and second-best performing models in each category, respectively.}
\centering
\footnotesize
\setlength{\tabcolsep}{10pt}
\renewcommand{\arraystretch}{1.1}
\resizebox{\linewidth}{!}
{
    \begin{tabular}{l|c|cccccc|c}
    \toprule
    \multirow{2}{*}{\raisebox{-0.5ex}{\textbf{Methods}}}
    & \multirow{2}{*}{\makecell{\textbf{Video}\\\textbf{Input}}}
    & \multicolumn{6}{c|}{\textbf{SQA3D (test)}}
    & \multirow{2}{*}{\raisebox{-0.5ex}{\textbf{Avg.}}}
    \\
    \cmidrule(lr){3-8}
    & ~
    & What & Is & How & Can & Which & Others\\
    \midrule 
    \rowcolor{navyblue!5}
    \multicolumn{9}{l}{\textcolor{black}{\textit{Task-Specific Models}}} \\
    SQA3D~\cite{ma2022sqa3d}     & \textcolor{Red}{\ding{55}} & \underline{31.6} & \textbf{63.8} & \textbf{46.0} & \underline{69.5} & \underline{43.9} & \underline{45.3} & \underline{46.6} \\
    3D-VisTA~\cite{zhu20233d}   & \textcolor{Red}{\ding{55}} & \textbf{34.8} & \underline{63.3} & \underline{45.4} & \textbf{69.8} & \textbf{47.2} & \textbf{48.1} & \textbf{48.5} \\
    \midrule
    \rowcolor{navyblue!5}
    \multicolumn{9}{l}{\textcolor{black}{\textit{3D/2.5D-Input Models}}} \\
    ChatScene~\cite{huang2024chatscene} & \textcolor{Red}{\ding{55}} & 45.4 & 67.0 & 52.0 & 69.5 & 49.9 & 55.0 & 54.6 \\
    LLaVA-3D~\cite{zhu2024llava}        & \textcolor{Red}{\ding{55}} & - & - & - & - & - & - & 55.6 \\
    Video-3D LLM~\cite{zheng2025video}  & \textcolor{Red}{\ding{55}} & 51.1 & 72.4 & 55.5 & 69.8 & \underline{51.3} & 56.0& 58.6 \\
    3DRS~\cite{huang2025mllms}        & \textcolor{Red}{\ding{55}} & \underline{54.4} & \underline{75.2} & \underline{57.0} & \textbf{72.2} & 49.9 & \underline{59.0}  & \underline{60.6} \\
    Ross3D~\cite{wang2025ross3d}       & \textcolor{Red}{\ding{55}} & \textbf{56.0} & \textbf{79.8} & \textbf{60.6} & \underline{70.4} & \textbf{55.3} & \textbf{60.1} & \textbf{63.0} \\
    \midrule 
    \rowcolor{navyblue!5}
    \multicolumn{9}{l}{\textcolor{black}{\textit{Video-Input Models}}} \\
    Qwen2-VL-7B~\cite{bai2025qwen2}          & \textcolor{Green}{\ding{51}} & 39.7 & 56.6 & 41.1 & 55.9 & 47.6 & 47.2& 46.5 \\
    Qwen2-VL-72B~\cite{bai2025qwen2}        & \textcolor{Green}{\ding{51}} & 41.7 & 56.3 & 41.5 & 55.6 & 44.5 & 48.0 & 47.0 \\
    LLaVA-Video-7B~\cite{zhang2024llavavideo}  & \textcolor{Green}{\ding{51}}& 42.7 & 56.3 & 47.5 & 55.3 & 50.1 & 47.2 & 48.5 \\
    Spatial-MLLM-4B~\cite{wu2025spatialmllm}  & \textcolor{Green}{\ding{51}} & \underline{45.9} & \underline{71.6} & \underline{55.1} & \textbf{69.5} & \textbf{52.0} & \underline{53.0} & \underline{55.9} \\
    \rowcolor{gray!10}
    \textbf{VLM$^2$-7B (Ours)}              & \textcolor{Green}{\ding{51}} & \textbf{54.5} & \textbf{74.8} & \textbf{58.1} & \underline{68.1} & \underline{51.6} & \textbf{58.7}  & \textbf{60.4} \\
    \bottomrule
    \end{tabular}
}
\label{tab:fullsqa3d}
\vspace{-1.5em}
\end{table*}

Table~\ref{tab:fullscanqa} and Table~\ref{tab:fullsqa3d} present additional evaluation results on the ScanQA~\cite{azuma2022scanqa} and SQA3D~\cite{ma2022sqa3d} benchmarks, complementing results reported in the main paper.

\noindent
\textbf{ScanQA.}
VLM$^2$ shows strong performance on ScanQA, consistently outperforming all video-input models, including LLaVA-Video-7B~\cite{zhang2024llavavideo} and Spatial-MLLM~\cite{wu2025spatialmllm}.
Although our method only takes video frames as input, it remains highly competitive against 3D/2.5D-input models that explicitly leverage depth or point clouds.
These results indicate that our 3D-aware representation provides effective geometric cues for 3D scene understanding.

\noindent
\textbf{SQA3D.}
We also report SQA3D results broken down by six question types (\textit{What}, \textit{Is}, \textit{How}, \textit{Can}, \textit{Which}, \textit{Others}).
As shown in Table~\ref{tab:fullsqa3d}, VLM$^2$ performs strongly among video-input models and remains competitive against methods that leverage additional 3D or 2.5D inputs. 
While Ross3D~\cite{wang2025ross3d} achieves higher overall performance than ours, it exploits extra point clouds to render BEV images for reconstructive supervision. 
Our method surpasses other 3D/2.5D-input models such as Video-3D LLM~\cite{zheng2025video} and 3DRS~\cite{huang2025mllms}, demonstrating solid overall performance on SQA3D.

\noindent Overall, these supplementary results show that VLM$^2$ is effective not only on video-based spatial reasoning benchmarks but also on 3D scene understanding tasks such as ScanQA and SQA3D, achieving strong performance across a broad range of evaluation settings and benchmarks.

\begin{table*}[!t]
\caption{
\textbf{Spatial reasoning performance on VSI-Bench~\cite{yang2025thinking} and VSTI-Bench~\cite{fan2025vlm3r} across varying video durations.}
Videos are grouped into Short ($<1$ min), Mid (1--2 min), and Long ($>2$ min).
For Spatial-MLLM~\cite{wu2025spatialmllm}, the original setting uses 16 input frames for evaluation,
while methods marked with $^*$ correspond to a 32-frame input setting, used to ensure a consistent frame setting across methods.
}
\centering
\footnotesize
\setlength{\tabcolsep}{4pt}
\resizebox{0.85\linewidth}{!}{
\begin{tabular}{l|cccc|cccc}
\toprule
\multirow{2}{*}{\raisebox{-0.75ex}{\textbf{Methods}}} &
\multicolumn{4}{c|}{\textbf{VSI-Bench}} &
\multicolumn{4}{c}{\textbf{VSTI-Bench}} \\
\cmidrule(lr){2-5}\cmidrule(lr){6-9} &
Avg. & Short & Mid & Long & Avg. & Short & Mid & Long \\
\midrule
\rowcolor{navyblue!5}
\multicolumn{9}{l}{\textcolor{black}{\textit{Open-sourced VLMs}}} \\
LongVA-7B~\cite{zhang2024long} & 31.1 & 33.5 & 32.2 & 28.8 & 38.2 & 36.9 & 38.4 & 40.5 \\
InternVL2-8B~\cite{chen2024far} & 34.6 & 35.8 & 33.8 & 34.9 & 49.1 & 46.2 & 49.9 & 53.3 \\
InternVL2-40B~\cite{chen2024far} & 36.1 & 38.5 & 36.0 & 35.1 & 50.3 & 46.0 & 51.6 & 56.2 \\
LongVILA-8B~\cite{xue2024longvila} & 21.2 & 20.1 & 21.0 & 21.9 & 34.3 & 34.6 & 33.9 & 34.7 \\
VILA-1.5-8B~\cite{lin2024vila} & 30.7 & 35.1 & 31.9 & 27.4 & 40.9 & 39.8 & 42.1 & 40.6 \\
VILA-1.5-40B~\cite{lin2024vila} & 31.6 & 33.3 & 32.3 & 30.1 & 44.8 & 41.9 & 45.1 & 50.0 \\
Qwen2.5VL-7B~\cite{bai2025qwen2} & 29.5 & 34.3 & 28.3 & 28.4 & 45.9 & 43.3 & 48.4 & 45.5 \\
Qwen2.5VL-72B~\cite{bai2025qwen2} & 36.3 & 39.2 & 36.6 & 34.7 & 49.4 & 45.1 & 50.8 & 55.2 \\
LLaVA-OneVision-7B~\cite{li2025llavaonevision} & 34.1 & 38.2 & 34.3 & 32.0 & 46.6 & 44.1 & 47.6 & 49.8 \\
LLaVA-OneVision-72B~\cite{li2025llavaonevision} & 41.2 & 43.0 & 41.5 & 40.2 & 52.2 & 48.2 & 52.9 & 59.2 \\
LLaVA-NeXT-Video-7B~\cite{llavanext} & 36.8 & 38.5 & 37.3 & 35.5 & 46.4 & 43.0 & 47.9 & 50.1 \\
LLaVA-NeXT-Video-72B~\cite{llavanext} & 41.1 & 41.3 & 40.7 & 41.4 & 51.3 & 47.6 & 51.9 & 57.6 \\
\midrule
\rowcolor{navyblue!5}
\multicolumn{9}{l}{\textcolor{black}{\textit{Spatial-Enhanced Models}}} \\
Spatial-MLLM-4B~\cite{wu2025spatialmllm} & 48.1 & 49.8 & 48.3 & 47.2 & - & - & - & - \\
Spatial-MLLM-4B$^*$~\cite{wu2025spatialmllm} & 49.0 & 49.7 & 49.3 & 48.5 & - & - & - & - \\
VG-LLM-4B~\cite{zheng2025vgllm} & 47.4 & 49.1 & 48.7 & 45.2 & - & - & - & - \\
VG-LLM-8B~\cite{zheng2025vgllm} & 50.5 & 51.9 & 49.7 & 50.6 & - & - & - & - \\
VLM-3R-7B~\cite{fan2025vlm3r} & \underline{63.2} & \underline{67.1} & \underline{64.7} & \underline{60.0} & \underline{63.6} & \underline{61.1} & \underline{64.7} & \underline{66.4} \\
\textbf{VLM$^2$-7B (Ours)} & \textbf{69.4} & \textbf{71.0} & \textbf{70.0} & \textbf{68.1} & \textbf{69.5} & \textbf{65.9} & \textbf{70.6} & \textbf{74.6} \\
\midrule
\rowcolor{gray!10}
\textit{Improve} $\uparrow$
& \textcolor{ForestGreen}{+6.2}
& \cellcolor{oai-green-200}{+3.9}
& \cellcolor{oai-green-400}{+5.3}
& \cellcolor{oai-green-600}{+8.1}
& \textcolor{ForestGreen}{+5.9}
& \cellcolor{oai-green-200}{+4.8}
& \cellcolor{oai-green-400}{+5.9}
& \cellcolor{oai-green-600}{+8.2}
\\
\bottomrule
\end{tabular}
}
\label{tab:long_horizon_appendix}
\vspace{-2.5em}
\end{table*}

\subsection{Additional Results Across Video Durations}
Table~\ref{tab:long_horizon_appendix} reports additional results across different video durations on VSI-Bench~\cite{yang2025thinking} and VSTI-Bench~\cite{fan2025vlm3r}, complementing the comparisons in the main paper.
We follow the same evaluation protocol and video-length partitioning as in Table 5 of the main paper, where videos are grouped into Short ($<1$ min), Mid (1--2 min), and Long ($>2$ min).
For Spatial-MLLM~\cite{wu2025spatialmllm}, the original setting uses 16 input frames for evaluation. We also include a 32-frame variant (marked with $*$) to ensure a consistent input setting across methods.

\subsection{Additional Ablations on Sem-Geo Fusion}
\begin{table}[!t]
\caption{\textbf{Additional ablations of Sem-Geo fusion strategies.}
In addition to \textit{Concat-MLP} and \textit{Cross-Attn} reported in the main paper, we further include the \textit{Add} variant under three 3D foundation models (VGGT~\cite{wang2025vggt}, CUT3R~\cite{wang2025cut3r}, $\pi^3$~\cite{wang2025pi}).
}
\centering
\setlength{\tabcolsep}{5pt}
\renewcommand{\arraystretch}{1.1}
\setlength\tabcolsep{3pt}
\resizebox{\columnwidth}{!}
{
\begin{tabular}{l|c|ccccc}
\toprule
\multirow{2}{*}{\raisebox{-0.5ex}{\textbf{Methods}}}  &
\multirow{2}{*}{\raisebox{-0.5ex}{\textbf{Avg.}}} &
\multicolumn{2}{c}{\textbf{Numerical Question}} &
\multicolumn{3}{c}{\textbf{Multiple-Choice Question}} \\
\cmidrule(lr){3-4}\cmidrule(lr){5-7} &
~ & Abs. Dist. & Room Size & Rel. Dist. & Rel. Dir. & Route Plan\\
\midrule
\rowcolor{navyblue!5}
\multicolumn{7}{l}{\textcolor{black}{\textit{Baseline}}} \\
LLaVA-NeXT-Video-SFT & 55.2 & 43.3 & 62.4 & 62.1 & 67.8 & 40.2 \\
\midrule
\rowcolor{navyblue!5}
\multicolumn{7}{l}{\textcolor{black}{\textit{Sem-Geo Fusion (Concat-MLP)}}} \\
Baseline + VGGT  & 41.4 & 38.6 & 45.6 & 38.7 & 47.6 & 36.6 \\
Baseline + CUT3R  & 43.3 & 39.8& 51.4 & 41.3 & 45.9 & 38.1 \\
Baseline + $\pi^3$  & 40.4 & 38.3 & 42.5 & 41.0 & 45.2 & 35.1\\
\midrule
\rowcolor{navyblue!5}
\multicolumn{7}{l}{\textcolor{black}{\textit{Sem-Geo Fusion (Cross-Attn)}}} \\
Baseline + VGGT   & 60.2 & \underline{51.0} & 63.9 & 61.4 & \underline{81.8} & 42.7 \\
Baseline + CUT3R  & 59.4 & 50.2 & 63.1 & 62.8 & 79.3 & 41.8 \\
Baseline + $\pi^3$ & \textbf{61.0} & \textbf{51.9} &  \textbf{64.7} & \underline{63.1} & \textbf{81.9} & \textbf{43.3} \\
\midrule
\rowcolor{navyblue!5}
\multicolumn{7}{l}{\textcolor{black}{\textit{Sem-Geo Fusion (Add)}}} \\
Baseline + VGGT   & 59.9 & 50.5 & 64.1 & 61.3 & 81.6 & 42.3 \\
Baseline + CUT3R  & 58.5 & 47.4 & \textbf{64.7} & 62.5 & 75.9 & 41.8 \\
Baseline + $\pi^3$ & \underline{60.7} & 50.9 & \underline{64.2} & \textbf{63.8} & 81.5 & \underline{42.9} \\
\bottomrule
\end{tabular}
}
\label{tab:ablation_sem_geo_appendix}
\vspace{-2.0em}
\end{table}

In Table 7 of the main paper, we compared two semantic-geometric (Sem-Geo) fusion strategies: \textit{Concat-MLP} and \textit{Cross-Attn}.
Here, we provide additional ablations by including the \textit{Add} fusion variant, which performs simple element-wise addition between semantic and geometric features.
Table~\ref{tab:ablation_sem_geo_appendix} reports the full comparison across all three fusion strategies.
We follow exactly the same training and evaluation setups as in the main experiments.
These supplementary results further confirm the consistent trends reported in the main paper:
$\pi^{3}$ consistently provides the strongest geometric prior, and \textit{Cross-Attn} remains the most effective Sem-Geo fusion strategy among all variants.

\section{Qualitative Results}
\begin{figure*}[t]
\centering
\begin{tcolorbox}[colback=blue!1!white, colframe=blue!5!white, width=\textwidth]
\begin{center}
    \includegraphics[width=1.0\textwidth]{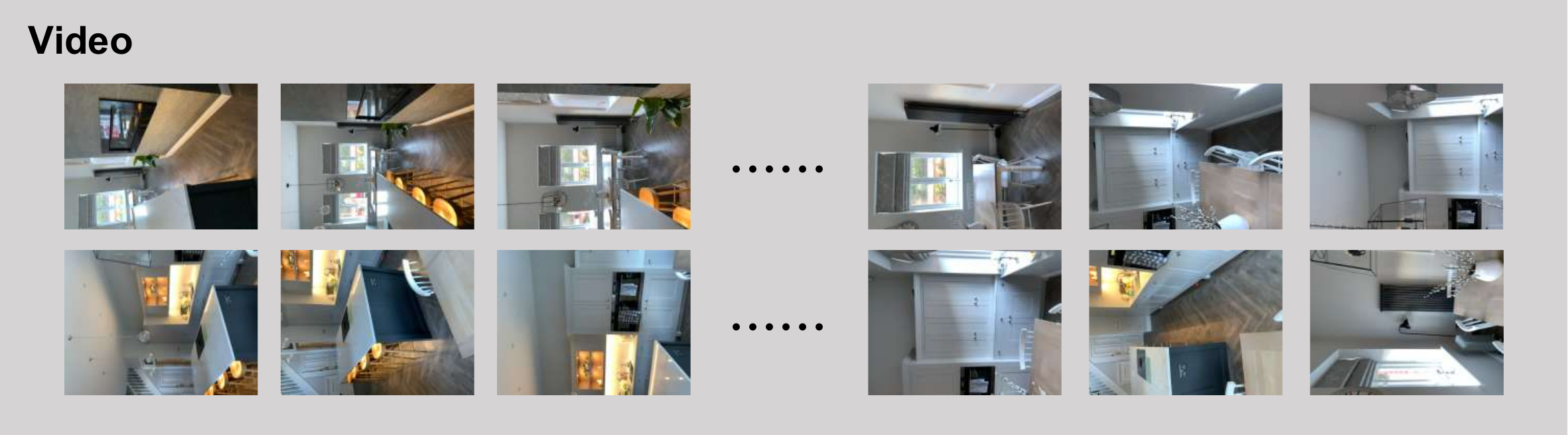}
\end{center}
\textbf{\textcolor{violet}{Question}}: \textit{How many chair(s) are in this room?} \\
\textbf{\textcolor{DarkBlue}{Answer}}: \textit{6} \\
\rule{\linewidth}{1.0pt}
\begin{center}
    \includegraphics[width=1.0\textwidth]{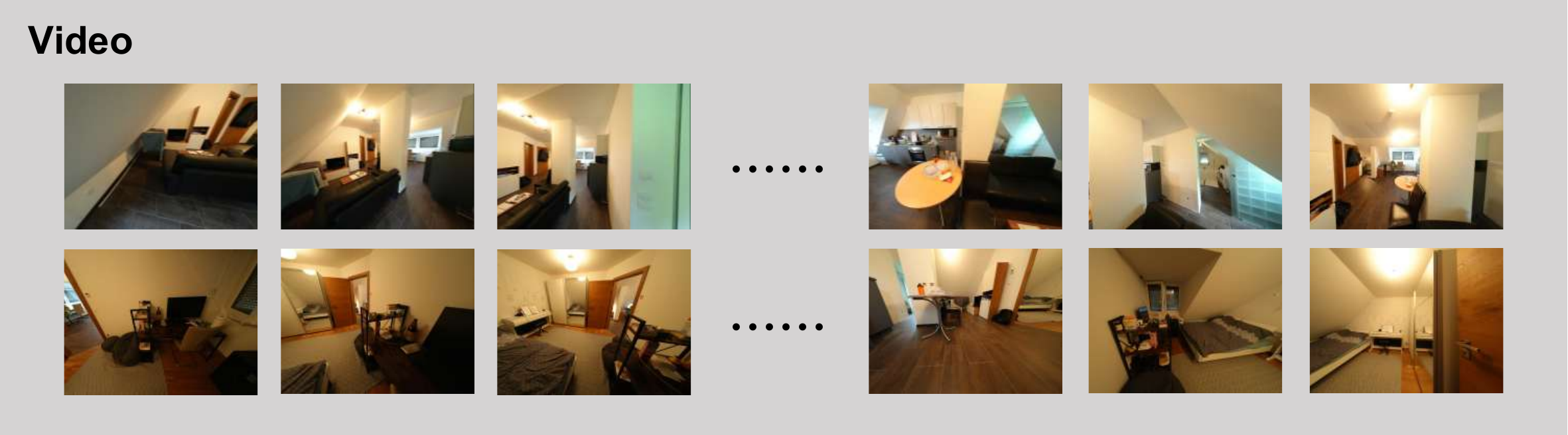}
\end{center}
\textbf{\textcolor{violet}{Question}}: \textit{Mearsuring from the cloest point of each object, what is the distance between the bed and the sofa (in meters)?} \\
\textbf{\textcolor{DarkBlue}{Answer}}: \textit{3.2} \\
\rule{\linewidth}{1.0pt}
\begin{center}
    \includegraphics[width=1.0\textwidth]{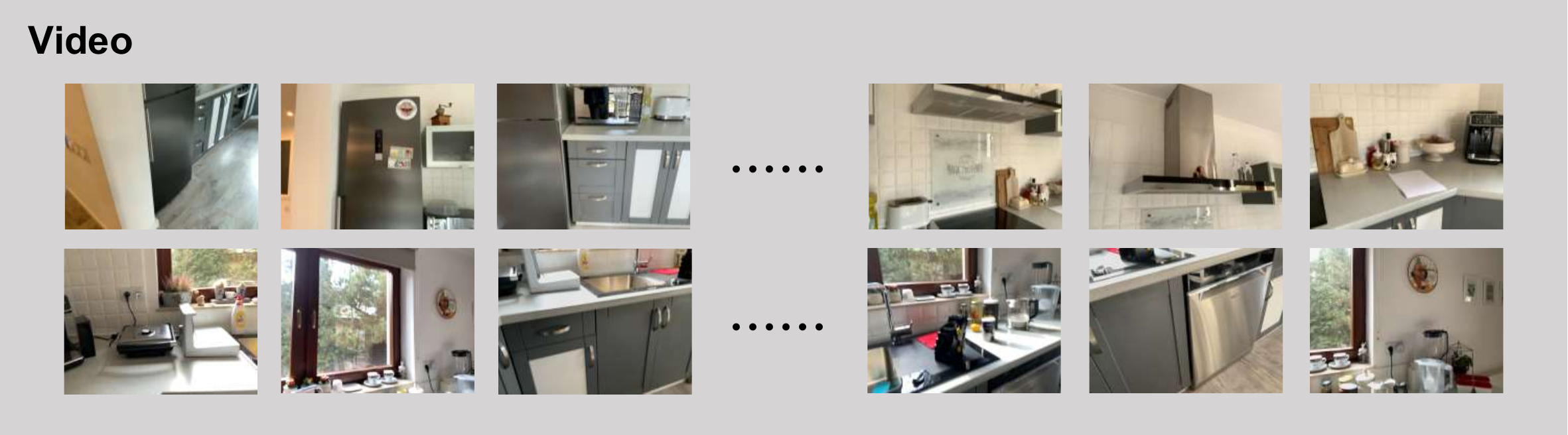}
\end{center}
\textbf{\textcolor{violet}{Question}}: \textit{You are a robot beginning at the kitchen sink and facing window. You want to navigate to the refrigerator. You will perform the following actions (Note: for each [please fill in], choose either `turn back,' `turn left,' or `turn right.'): 1. [please fill in] 2. Go forward until the refrigerator. You have reached the final destination. \\
Options: A. Turn Back B. Turn Right C. Turn Left} \\
\textbf{\textcolor{DarkBlue}{Answer}}: \textit{A} \\
\end{tcolorbox}
\caption{\textbf{Qualitative examples on VSI-Bench~\cite{yang2025thinking}.}}
\label{qual_vsi1}
\end{figure*}

\begin{figure*}[t!]
\centering
\begin{tcolorbox}[colback=blue!1!white, colframe=blue!5!white, width=\textwidth]
\begin{center}
    \includegraphics[width=1.0\textwidth]{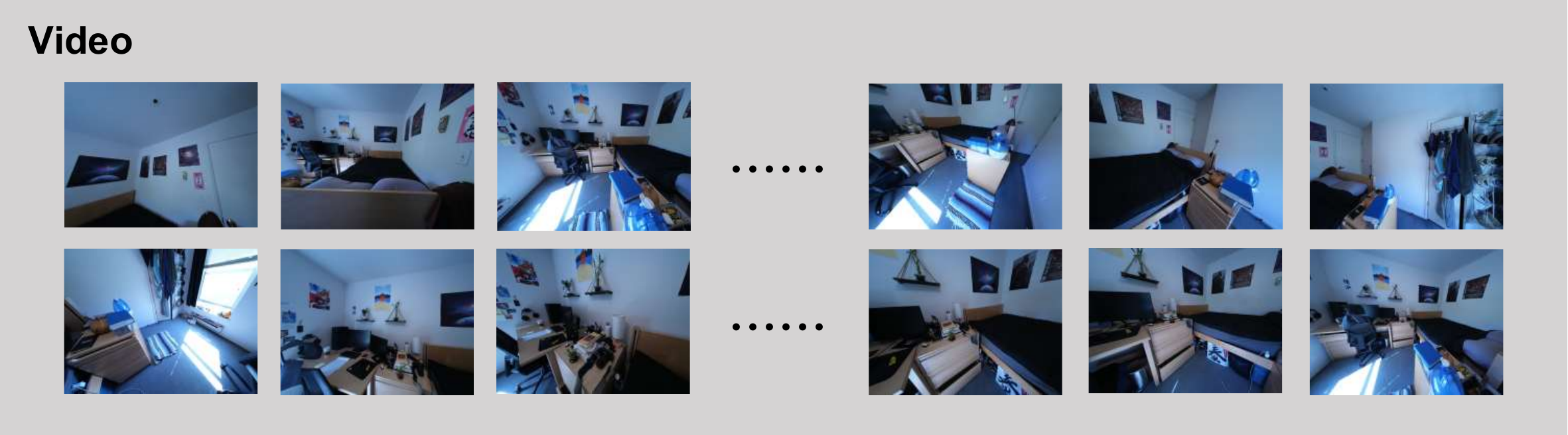}
\end{center}
\textbf{\textcolor{violet}{Question}}: \textit{What is the size of this room (in square meters)? If multiple rooms are shown, estimate the size of the combined space.} \\
\textbf{\textcolor{DarkBlue}{Answer}}: \textit{10.5} \\
\rule{\linewidth}{1.0pt}
\begin{center}
    \includegraphics[width=1.0\textwidth]{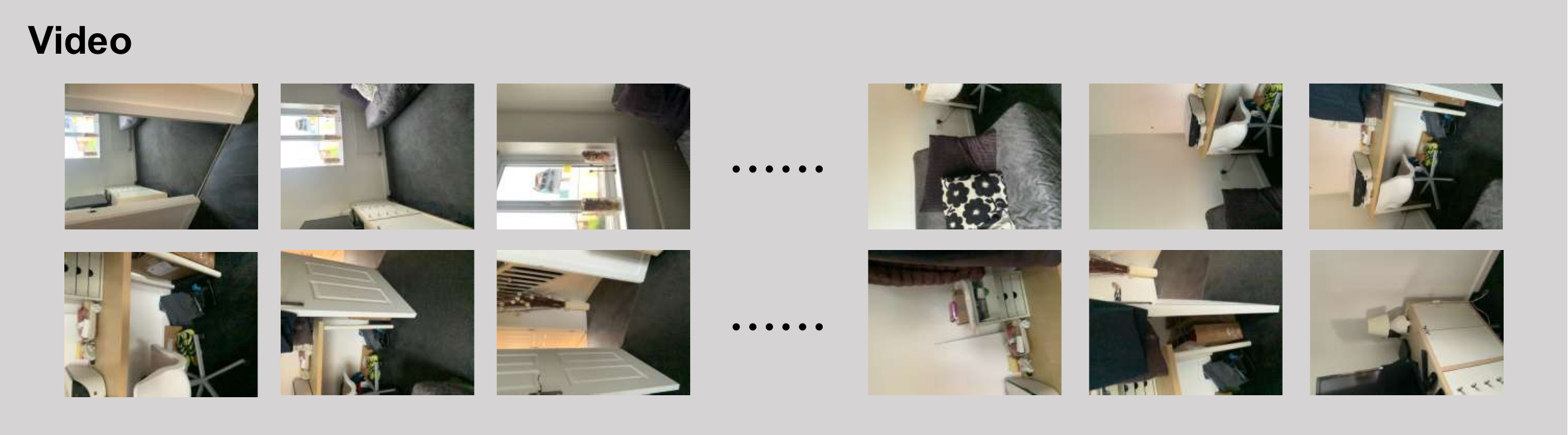}
\end{center}
\textbf{\textcolor{violet}{Question}}: \textit{What is the length of the longest dimension (length, width, or height) of the chair, measured in centimeters?} \\
\textbf{\textcolor{DarkBlue}{Answer}}: \textit{69} \\
\rule{\linewidth}{1.0pt}
\begin{center}
    \includegraphics[width=1.0\textwidth]{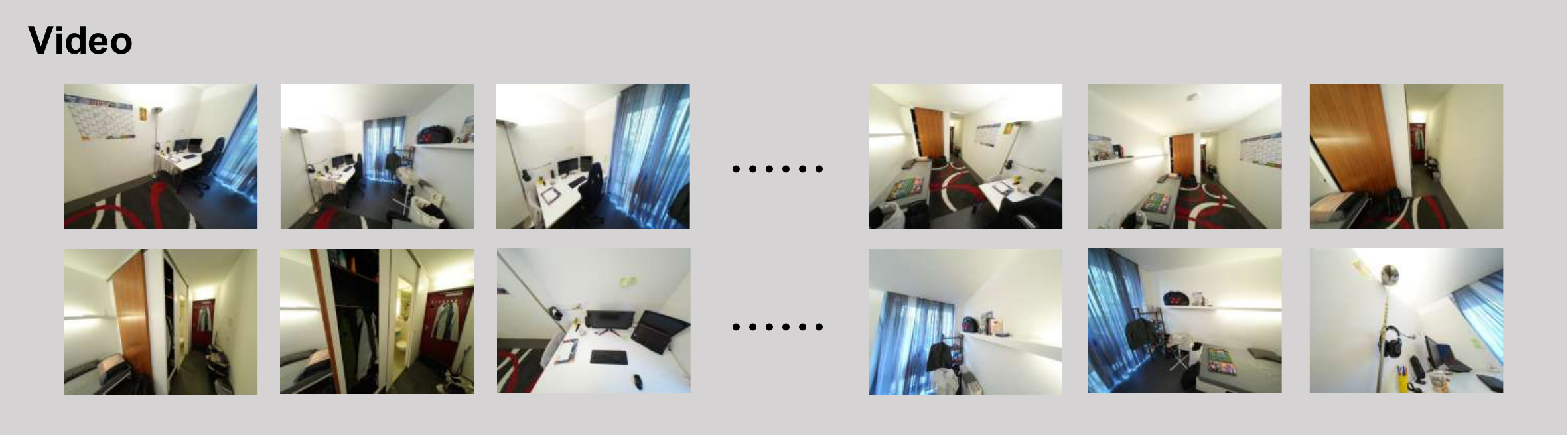}
\end{center}
\textbf{\textcolor{violet}{Question}}: \textit{What will be the first-time appearance order of the following categories in the video: basket, door, pillow, laptop? Options: A. basket, pillow, door, laptop B. pillow, door, laptop, basket C. basket, door, pillow, laptop D. door, basket, pillow, laptop} \\
\textbf{\textcolor{DarkBlue}{Answer}}: \textit{B} \\
\end{tcolorbox}
\caption{\textbf{Qualitative examples on VSI-Bench~\cite{yang2025thinking}.}}
\label{qual_vsi2}
\end{figure*}

\begin{figure*}[t]
\centering
\begin{tcolorbox}[colback=blue!1!white, colframe=blue!5!white, width=\textwidth]
\begin{center}
    \includegraphics[width=1.0\textwidth]{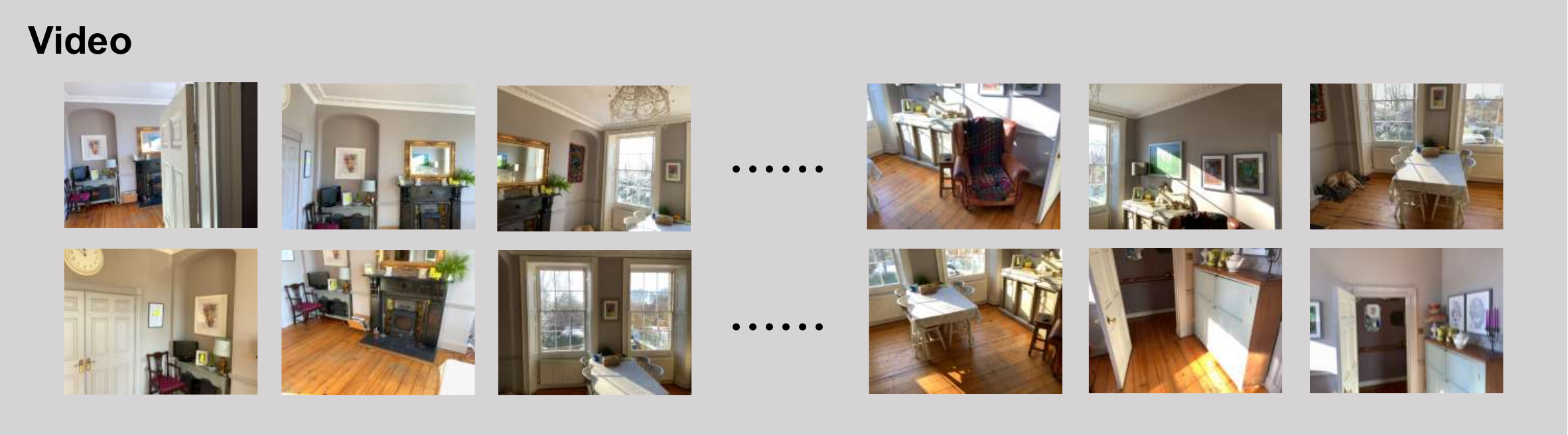}
\end{center}
\textbf{\textcolor{violet}{Question}}: \textit{Measuring from the closest point of each object, which of these objects (chair, table, tv, fireplace) is the cloest to the sofa? Options: A. chair B. table C. tv D. fireplace} \\
\textbf{\textcolor{DarkBlue}{Answer}}: \textit{B} \\
\rule{\linewidth}{1.0pt}
\begin{center}
    \includegraphics[width=1.0\textwidth]{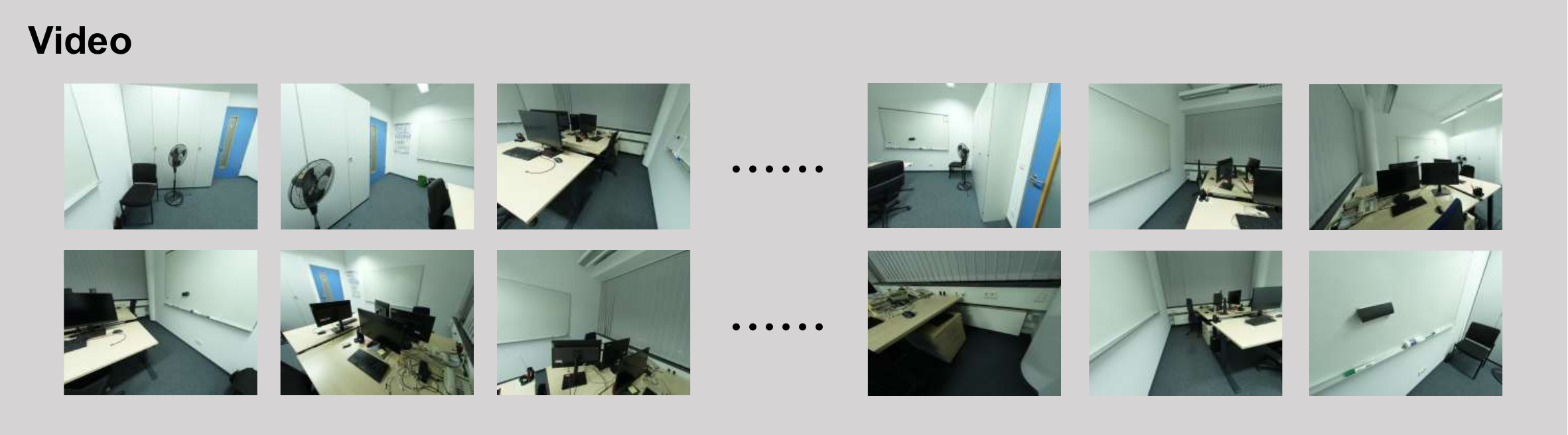}
\end{center}
\textbf{\textcolor{violet}{Question}}: \textit{If I am standing by the computer tower and facing the heater, is the door to my left, right, or back? An object is to my back if I would have to turn at least 135 degrees in order to face it. Options: A. back B. right C. left} \\
\textbf{\textcolor{DarkBlue}{Answer}}: \textit{A} \\
\rule{\linewidth}{1.0pt}
\begin{center}
    \includegraphics[width=1.0\textwidth]{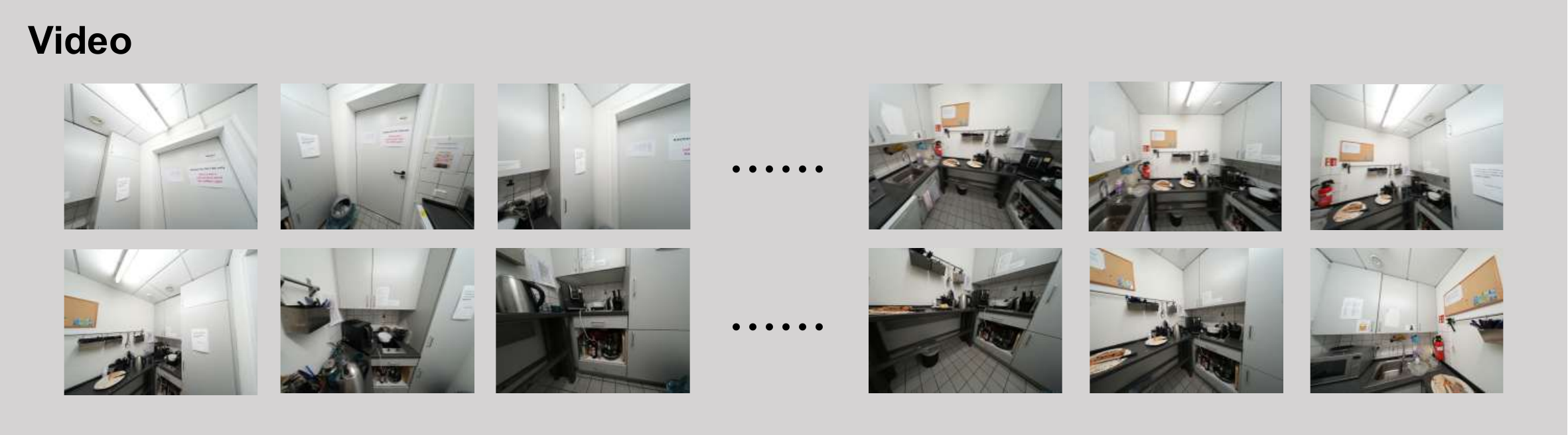}
\end{center}
\textbf{\textcolor{violet}{Question}}: \textit{If I am standing by the ceiling light and facing the door, is the kettle to my front-left, front-right, back-left, or back-right? The directions refer to the quadrants of a Cartesian plane (if I am standing at the origin and facing along the positive y-axis). Options: A. front-left B. back-right C. front-right D. back-left} \\
\textbf{\textcolor{DarkBlue}{Answer}}: \textit{D} \\
\end{tcolorbox}
\caption{\textbf{Qualitative examples on VSI-Bench~\cite{yang2025thinking}.}}
\label{qual_vsi3}
\end{figure*}

\begin{figure*}[t]
\centering
\begin{tcolorbox}[colback=blue!1!white, colframe=blue!5!white, width=\textwidth]
\begin{center}
    \includegraphics[width=1.0\textwidth]{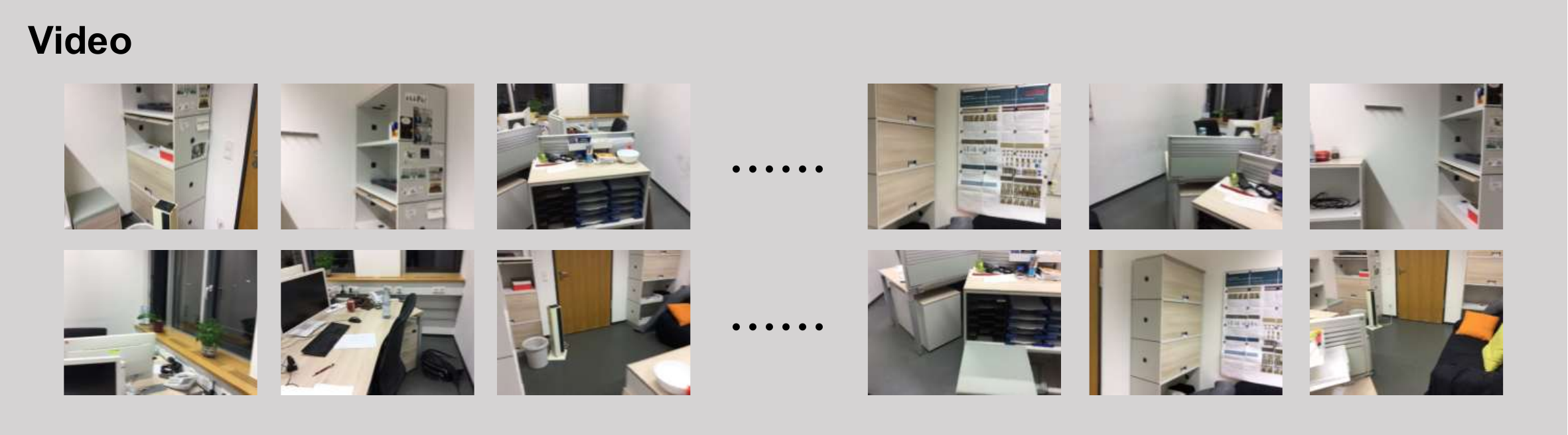}
\end{center}
\textbf{\textcolor{violet}{Question}}: \textit{Measuring from the closest point of each object, which of these objects (plant, window) is the closest to the camera in frame 4 of 32? Options: A. plant B. window} \\
\textbf{\textcolor{DarkBlue}{Answer}}: \textit{B} \\
\rule{\linewidth}{1.0pt}
\begin{center}
    \includegraphics[width=1.0\textwidth]{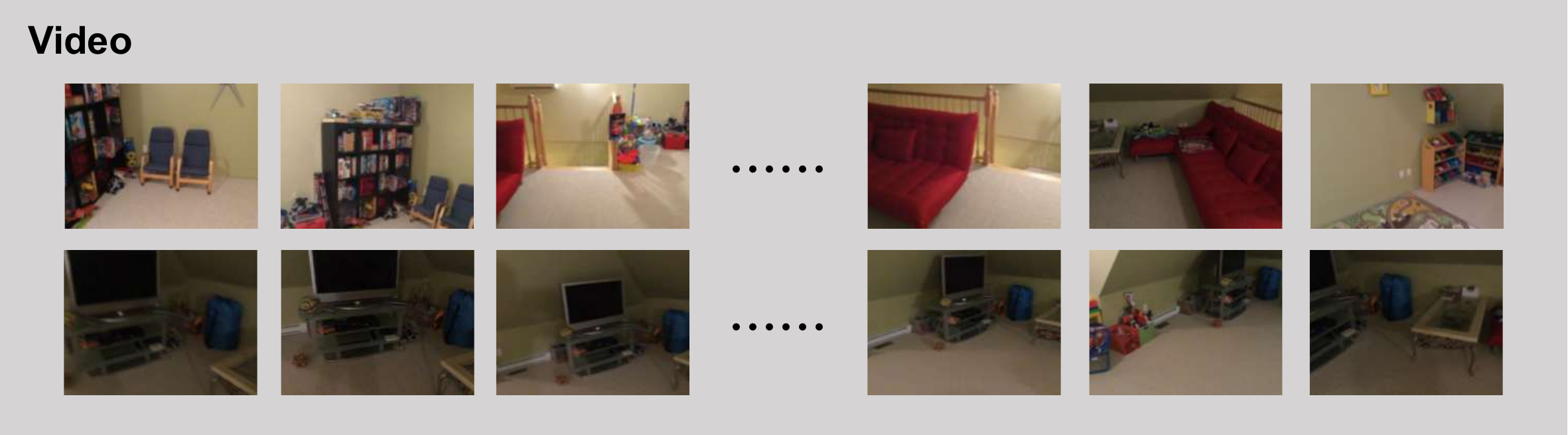}
\end{center}
\textbf{\textcolor{violet}{Question}}: \textit{During the sequence between frame 4 and frame 23 of 32, what was the primary consistent direction of the camera's movement relative to its orientation at the start? The options are Forward, Left, and Right. \\
Options: A. Forward B. Left C. Right} \\
\textbf{\textcolor{DarkBlue}{Answer}}: \textit{C} \\
\rule{\linewidth}{1.0pt}
\begin{center}
    \includegraphics[width=1.0\textwidth]{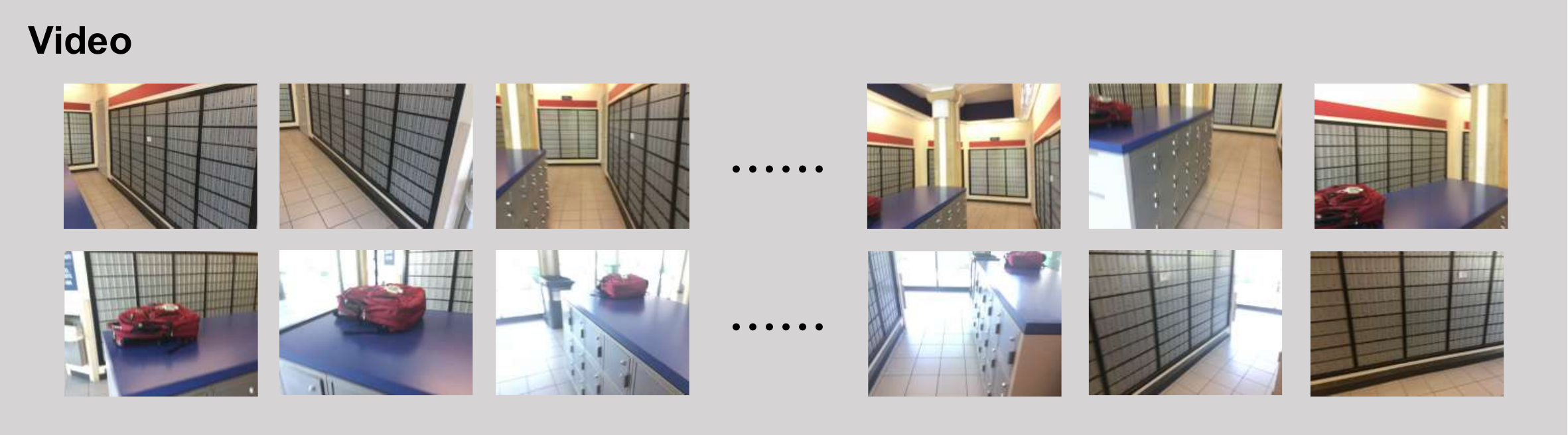}
\end{center}
\textbf{\textcolor{violet}{Question}}: \textit{What is the approximate distance (in meters) between the camera (or the person filming) and the nearest point of the backpack in frame 10 of 32?} \\
\textbf{\textcolor{DarkBlue}{Answer}}: \textit{0.7} \\
\end{tcolorbox}
\caption{\textbf{Qualitative examples on VSTI-Bench~\cite{fan2025vlm3r}.}}
\label{qual_vsti1}
\end{figure*}

\begin{figure*}[t]
\centering
\begin{tcolorbox}[colback=blue!1!white, colframe=blue!5!white, width=\textwidth]
\begin{center}
    \includegraphics[width=1.0\textwidth]{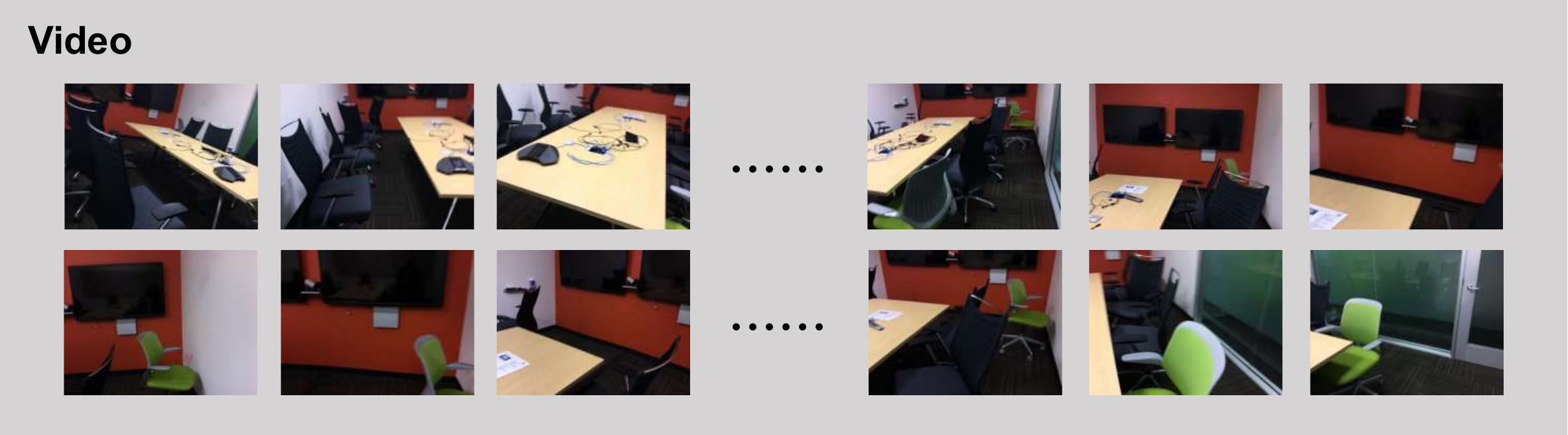}
\end{center}
\textbf{\textcolor{violet}{Question}}: \textit{Approximately how far (in meters) did the camera move between frame 24 and frame 31 of 32?} \\
\textbf{\textcolor{DarkBlue}{Answer}}: \textit{1.2} \\
\rule{\linewidth}{1.0pt}
\begin{center}
    \includegraphics[width=1.0\textwidth]{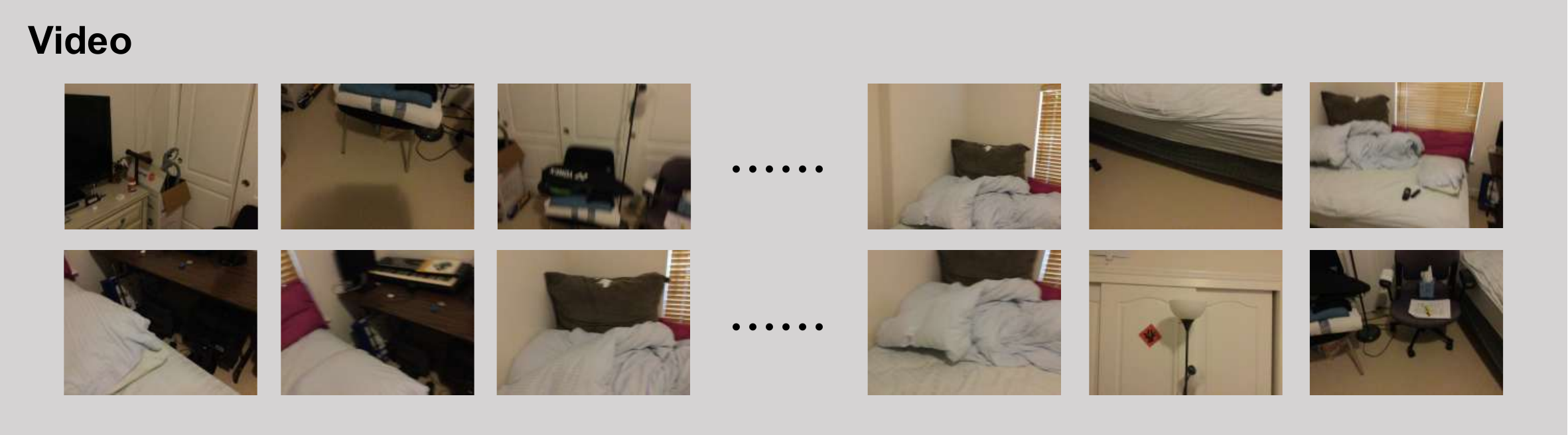}
\end{center}
\textbf{\textcolor{violet}{Question}}: \textit{In frame 13 of 31, relative to monitor, is bed to the [Left/Right]? Options: A. Right B. Left} \\
\textbf{\textcolor{DarkBlue}{Answer}}: \textit{B} \\
\rule{\linewidth}{1.0pt}
\begin{center}
    \includegraphics[width=1.0\textwidth]{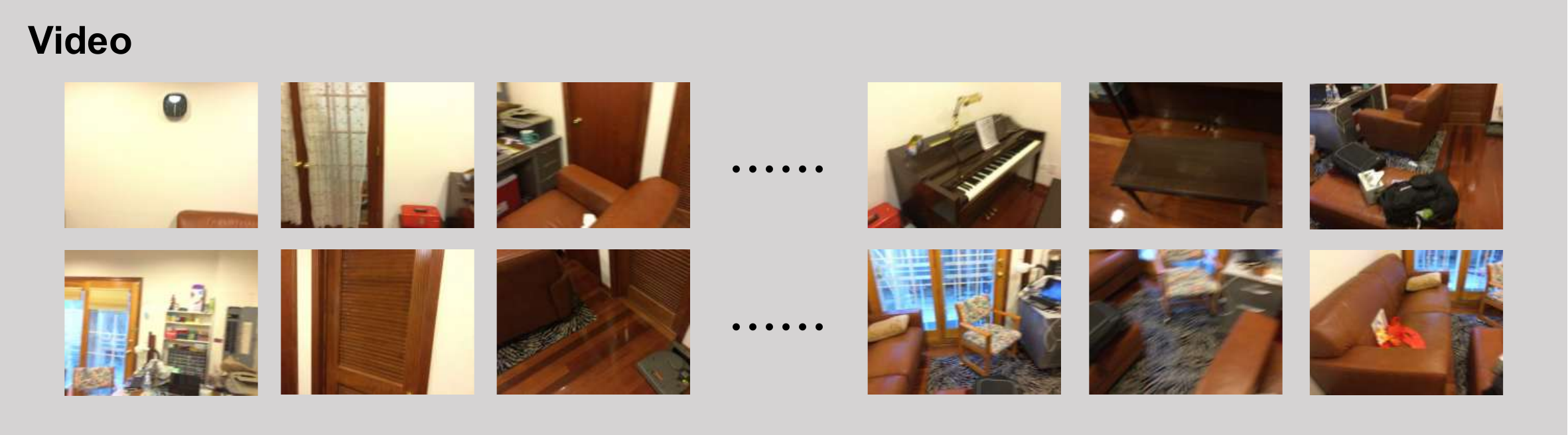}
\end{center}
\textbf{\textcolor{violet}{Question}}: \textit{In frame 1 of 31, relative to clock, is piano to the [Up/Down]? Options: A. Down B. Up} \\
\textbf{\textcolor{DarkBlue}{Answer}}: \textit{A} \\
\end{tcolorbox}
\caption{\textbf{Qualitative examples on VSTI-Bench~\cite{fan2025vlm3r}.}}
\label{qual_vsti2}
\end{figure*}

Figures~\ref{qual_vsi1} to ~\ref{qual_vsti2} show qualitative examples of VLM$^2$ on the VSI-Bench~\cite{yang2025thinking} and VSTI-Bench~\cite{fan2025vlm3r} benchmarks.
We include cases covering various spatial reasoning tasks, such as configurational estimation, measurement estimation, and temporal reasoning.
These visual examples illustrate the model's spatial reasoning capability across different tasks.

\end{document}